\documentclass{article}

\usepackage{microtype}
\usepackage{graphicx}
\usepackage{subcaption}
\usepackage{booktabs} 
\usepackage{multirow} 

\usepackage{hyperref}

\usepackage{amd}

\usepackage{amsmath}
\usepackage{amssymb}
\usepackage{mathtools}
\usepackage{amsthm}

\usepackage{graphicx}
\usepackage{booktabs}
\usepackage{caption}
\usepackage[table]{xcolor}
\usepackage{tcolorbox}
\usepackage{tabularx} 
\usepackage{array}
\usepackage[labelfont=bf,justification=centering]{caption}

\usepackage{microtype}
\usepackage{hyperref}
\usepackage{url}
\usepackage{booktabs}
\usepackage{multirow}
\usepackage[table]{xcolor}
\usepackage{arydshln}
\usepackage{tabularx}
 \usepackage{amsmath}
 \usepackage{tabularx,colortbl}
\usepackage{graphicx}
 \usepackage{xspace}
\usepackage{times}
\usepackage{latexsym}
\usepackage{soul}
\usepackage{tabularx,colortbl}
\usepackage{tabularx}
\usepackage{makecell}
\usepackage{multirow}
\usepackage{float}
\usepackage{arydshln} 
\usepackage{wrapfig}
\usepackage{graphics}
\usepackage{graphicx}
\usepackage{listings}
\usepackage{amssymb}
\usepackage{enumitem}
\usepackage{tcolorbox}
\usepackage{subcaption}
\usepackage{makecell}

 \newcommand{\ours}{{HyLo}\xspace}

\usepackage[capitalize,noabbrev]{cleveref}
\usepackage{tikz}
\usepackage{xspace}
\usepackage{xspace}

\newcommand{\AMDlogo}{\includegraphics[height=0.28in]{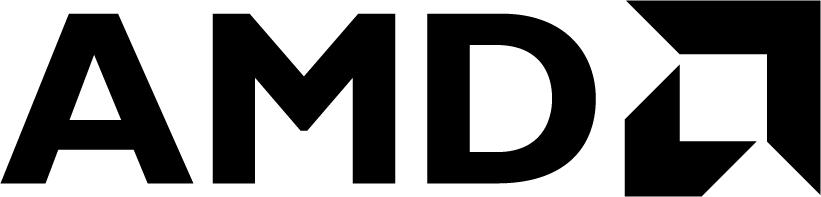}}

\theoremstyle{plain}

\theoremstyle{definition}

\theoremstyle{remark}

\usepackage[textsize=tiny]{todonotes}

\usepackage{lineno}

\definecolor{darkblue}{rgb}{0, 0, 0.5}
\hypersetup{colorlinks=true, citecolor=darkblue, linkcolor=darkblue, urlcolor=darkblue}
\definecolor{oursgray}{RGB}{245,245,245}
\definecolor{oursgray}{gray}{0.94}

\amdtitlerunning{Long-Context Aware Upcycling: A New Frontier for Hybrid LLM Scaling}

\begin{document}

\onecolumn

  \amdtitle{Long-Context Aware Upcycling: \\ A New Frontier for Hybrid LLM Scaling}



  \begin{amdauthorlist}
    \amdauthor{Parsa Ashrafi Fashi}{equal,comp} 
    \amdauthor{Utkarsh Saxena}{equal,comp}
    \amdauthor{Mehdi Rezagholizadeh}{equal,comp} 
    \amdauthor{Aref Jafari}{comp}
    \amdauthor{Akash Haridas}{comp}\\
    \amdauthor{Mingyu Yang}{comp}
    \amdauthor{Vansh Bhatia}{comp}
    \amdauthor{Guihong Li}{comp}
    \amdauthor{Vikram Appia}{comp}
    \amdauthor{Emad Barsoum}{comp}
  \end{amdauthorlist}

  \amdaffiliation{comp}{AMD}

\amdcorrespondingauthor{Correspondence}{\{parsa.fashi, utkarsh.saxena, mehdi.rezagholizadeh, aref.jafari\}@amd.com}


  \vskip 0.3in



\amdPrintAffiliationsInline
\amdEqualContribution

\begin{abstract}

Hybrid sequence models that combine efficient Transformer components with linear sequence modeling blocks are a promising alternative to pure Transformers, but most are still pretrained from scratch and therefore fail to reuse existing Transformer checkpoints. We study \emph{upcycling} as a practical path to convert pretrained Transformer LLMs into hybrid architectures while preserving short-context quality and improving long-context capability. We call our solution \emph{HyLo} (\underline{\textbf{HY}}brid \underline{\textbf{LO}}ng-context): a long-context upcycling recipe that combines architectural adaptation with efficient Transformer blocks, Multi-Head Latent Attention (MLA), and linear blocks (Mamba2 or Gated DeltaNet), together with staged long-context training and teacher-guided distillation for stable optimization. HyLo extends usable context length by up to $32\times$ through efficient post-training and reduces KV-cache memory by more than $90\%$, enabling up to 2M-token prefill and decoding in our \texttt{vLLM} inference stack, while comparable Llama baselines run out of memory beyond 64K context. Across 1B- and 3B-scale settings (Llama- and Qwen-based variants), HyLo delivers consistently strong short- and long-context performance and significantly outperforms state-of-the-art upcycled hybrid baselines on long-context evaluations such as RULER. Notably, at similar scale, HyLo-Qwen-1.7B trained on only 10B tokens significantly outperforms JetNemotron (trained on 400B tokens) on GSM8K, Lm-Harness common sense reasoning and RULER-64K.   

\end{abstract}

\section{Introduction}
Transformer-based large language models (LLMs) have achieved remarkable success across a broad spectrum of tasks, including natural language understanding, reasoning, and code generation \citep{vaswani2017attention, brown2020language, chowdhery2022palm}. These advances have been driven by scaling both model size and training data, resulting in state-of-the-art performance but at the cost of substantial computational and financial resources. Consequently, training new models from scratch has become increasingly prohibitive, motivating the search for more efficient architectures and training paradigms.

Recently, hybrid architectures that combine attention mechanisms with more efficient sequence modeling components such as state space models or linear attention have emerged as a promising direction. These models aim to retain the expressive power of Transformers while improving computational efficiency, particularly for long sequences. Notable examples include Jamba \citep{jamba}, Samba \citep{ren2024samba}, Qwen3-Next \citep{qwen3next2025}, and Kimi-Linear \citep{team2025kimi}, which demonstrate competitive performance with improved efficiency. However, these approaches largely rely on training from scratch, effectively replicating the immense cost associated with developing Transformer-based LLMs.

To address this limitation, a growing line of work explores model upcycling, which seeks to convert existing pre-trained Transformer models into hybrid architectures without discarding their learned knowledge. Instead of training a hybrid model from scratch, upcycling methods reuse the parameters of a pre-trained Transformer and transform its architecture, followed by continued training. The central goal is to enable efficient knowledge transfer from a source model to a target hybrid model, thereby reducing training cost while maintaining performance.
\begin{wrapfigure}{r}{0.5\textwidth}
    \centering
    \includegraphics[width=0.5\textwidth]{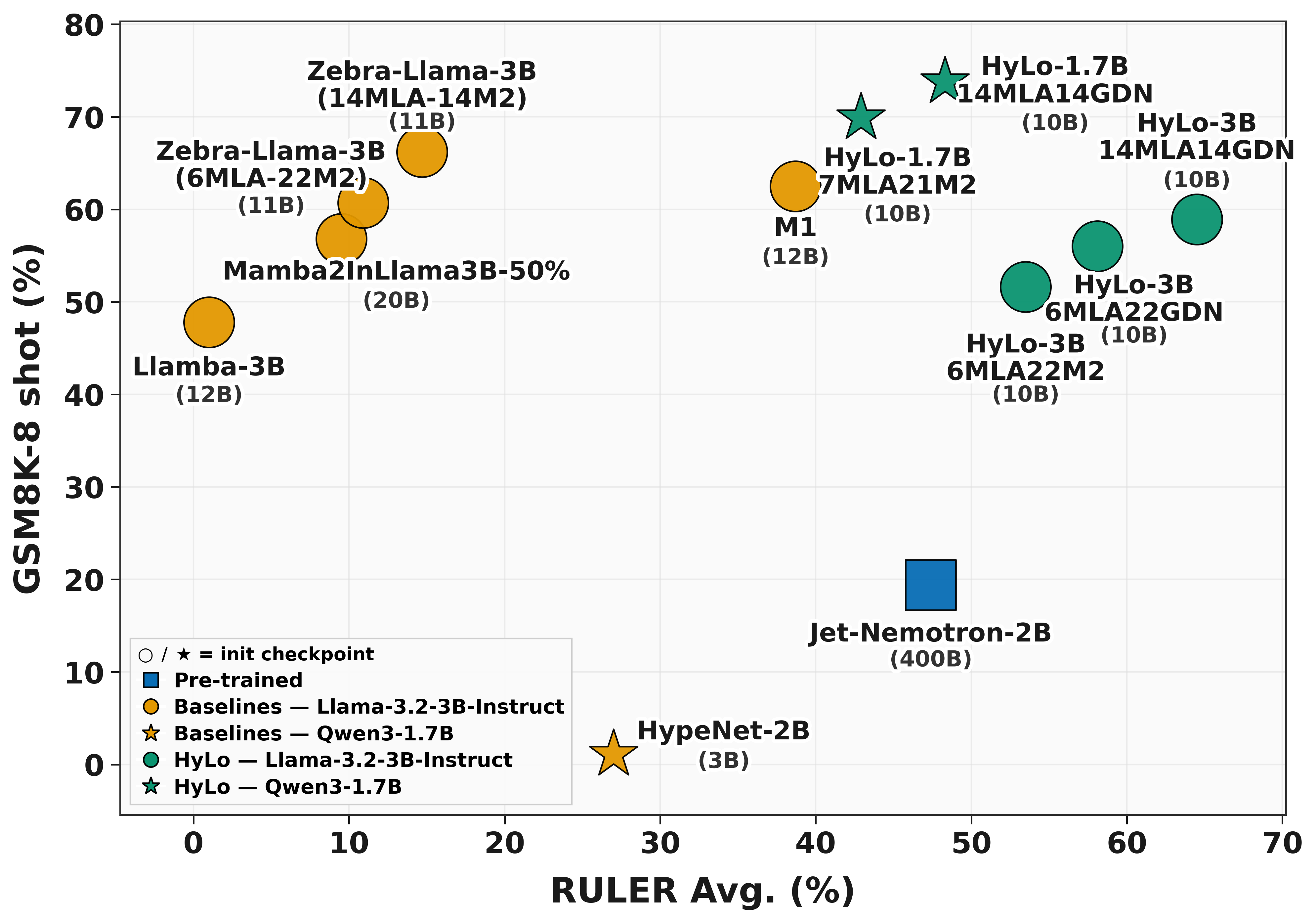}
    \caption{Short-context math performance and average RULER accuracy across 8K, 16K, 32K and 64K context lengths. HyLo models achieve competitive short context performance while outperforming baselines on long-context benchmark in a limited upcycling data budget.}
    \label{fig:fig1}
\end{wrapfigure}
 
Several recent works have proposed different upcycling approaches, including MambaInLlama \citep{wang2024mambainllama}, Mohawk \citep{bick2024transformers}, Lamba \citep{bick2025llamba}, and Zebra Llama \citep{yang2025zebra}. These methods provide initial evidence that it is possible to re-purpose Transformer models into hybrid architectures while preserving accuracy to a certain extent.

\begin{figure}[t]
    \centering
    \begin{minipage}[t]{0.25\columnwidth}
        \centering
        \includegraphics[width=\linewidth]{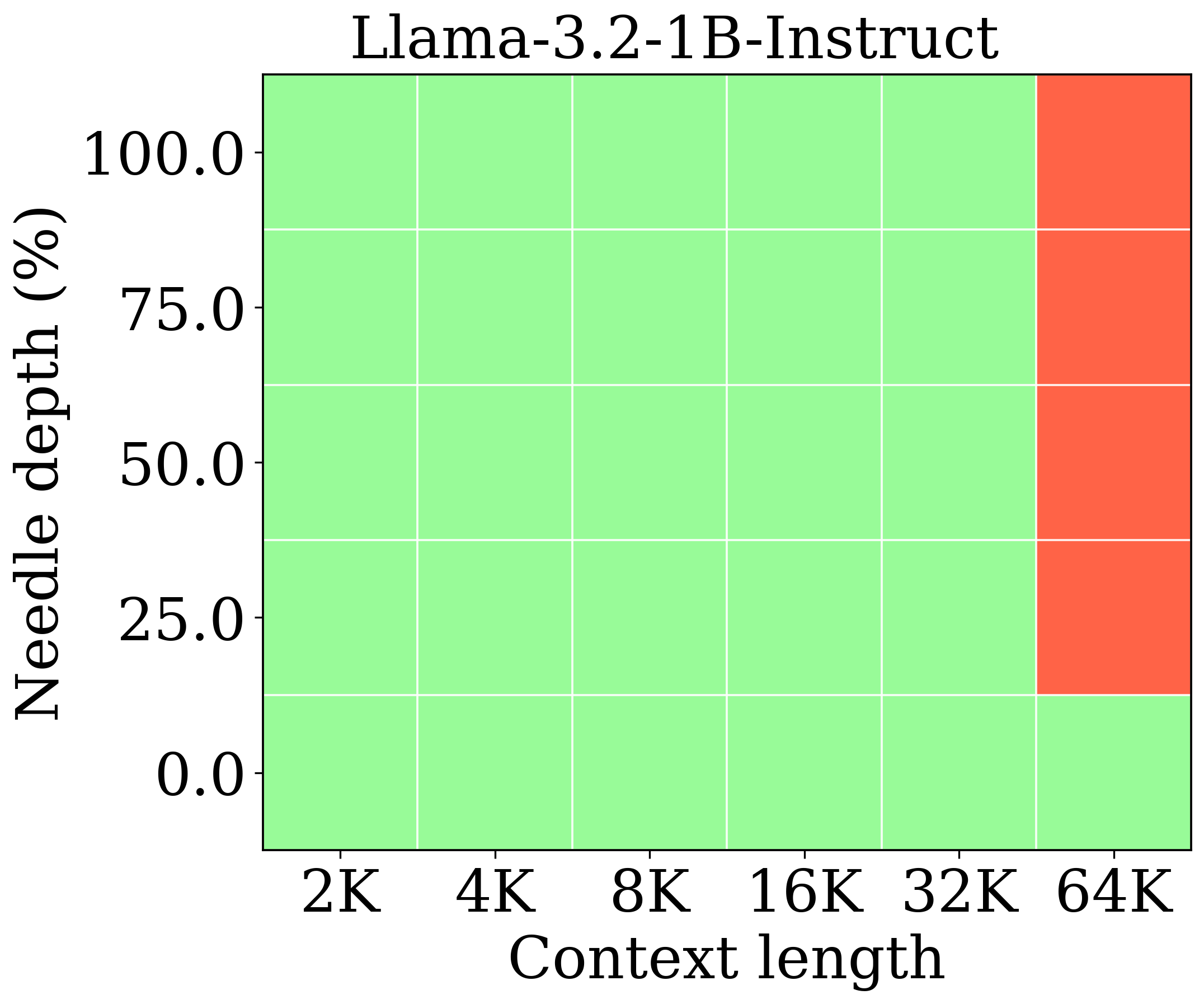}
    \end{minipage}\hfill
    \begin{minipage}[t]{0.25\columnwidth}
        \centering
        \includegraphics[width=\linewidth]{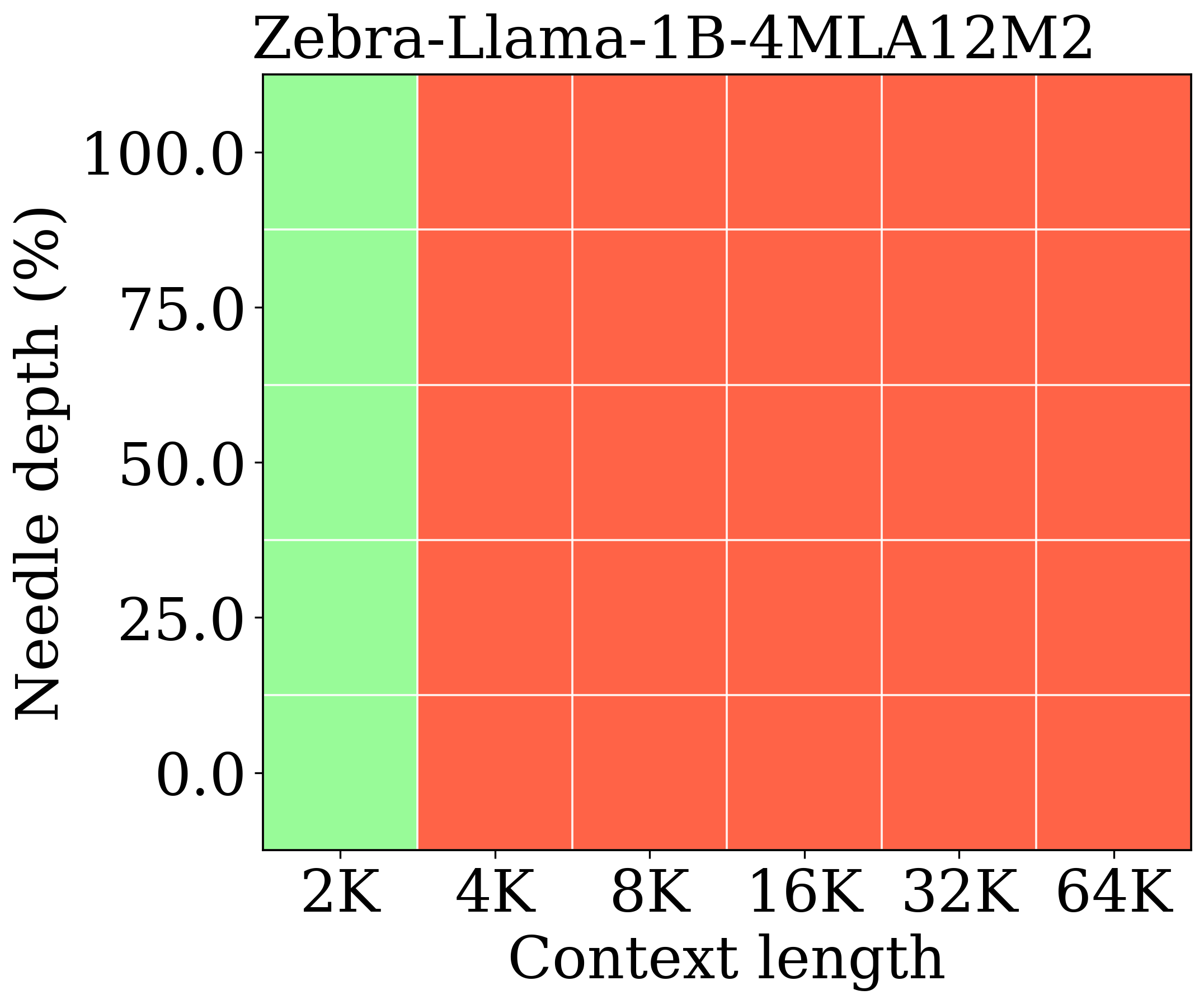}
    \end{minipage}\hfill
    \begin{minipage}[t]{0.25\columnwidth}
        \centering
        \includegraphics[width=\linewidth]{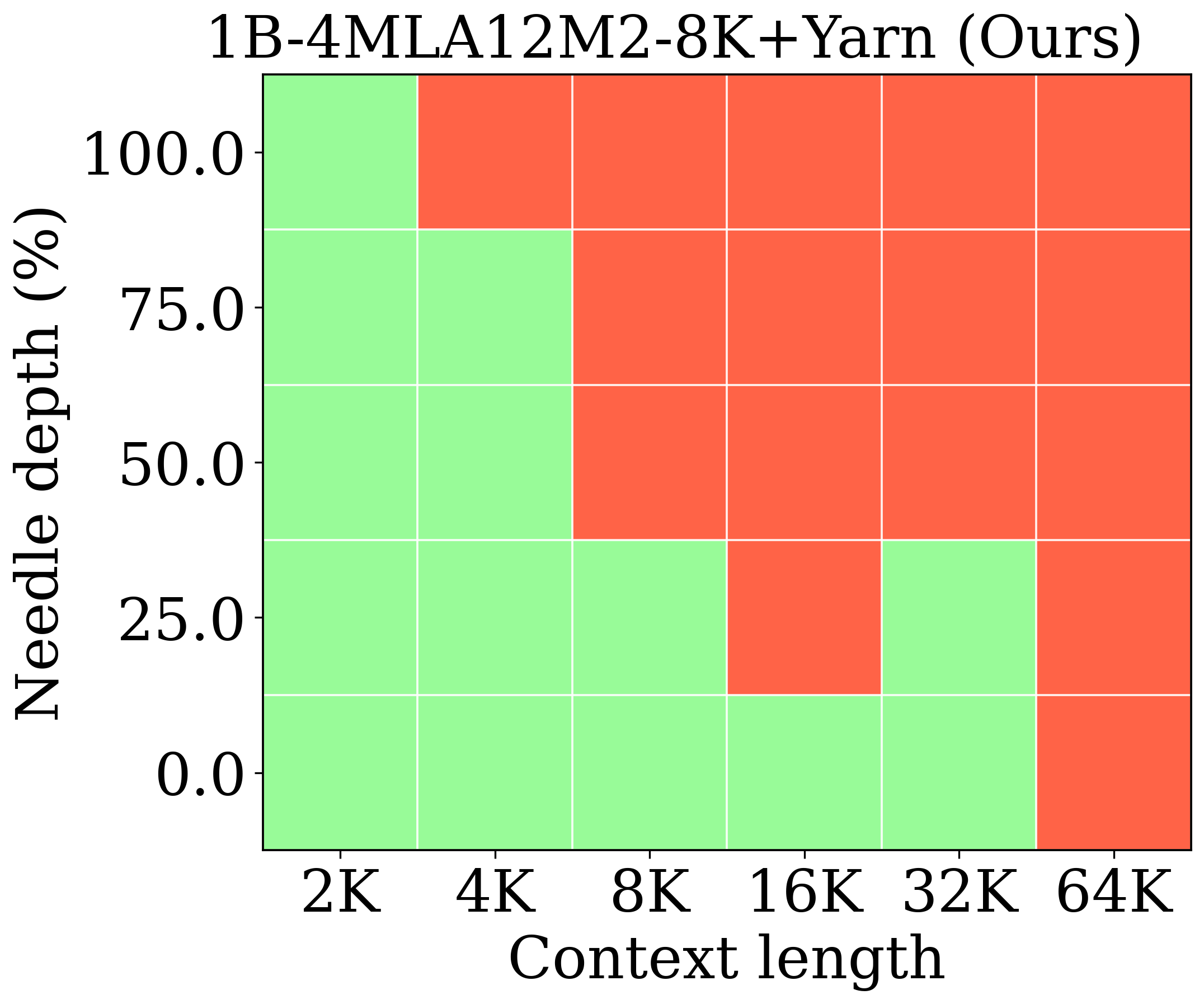}
    \end{minipage}\hfill
    \begin{minipage}[t]{0.25\columnwidth}
        \centering
        \includegraphics[width=\linewidth]{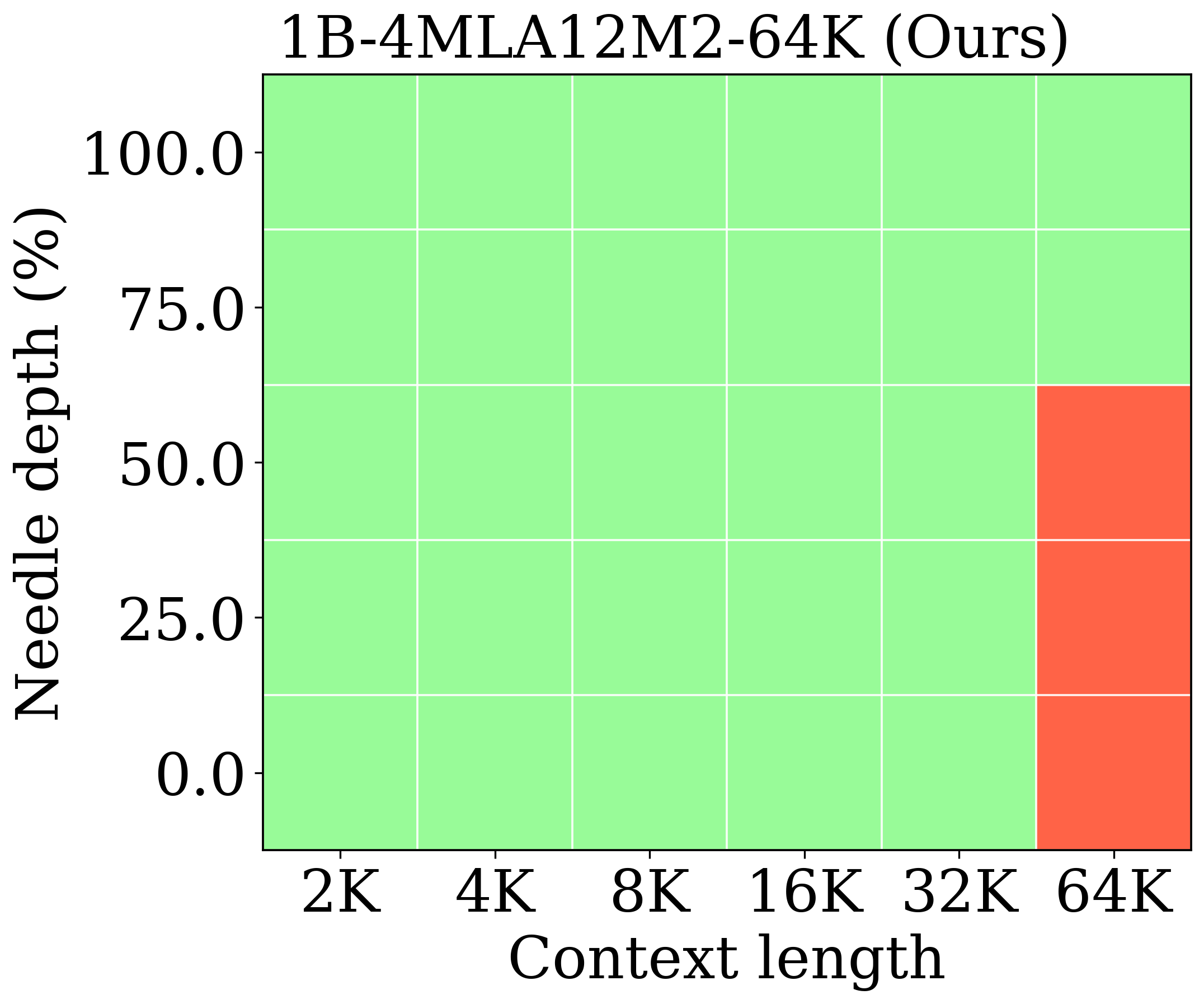}
    \end{minipage}
    \caption{Evaluation on synthetic needle in haystack benchmark demonstrates that our upcycled hybrid 4MLA12M2 model (at only $3.9\%$ KV cache footprint) achieves comparable performance to Llama-3.2-1B and surpasses Zebra-Llama. Furthermore, finetuning at 64K sequence length surpasses performance compared to 8K sequence length showcasing the need for long context finetuning.}
    \label{fig:niah_synthetic}
\end{figure}

However, existing upcycling approaches predominantly focus on maintaining short-context performance metrics such as perplexity or benchmark accuracy. In doing so, they largely overlook the long-context ability of modern LLMs which  has become increasingly important for real-world applications, including document understanding, code completion, and multi-hop reasoning. While hybrid architectures are often motivated by their theoretical advantages in handling long sequences, it remains unclear whether upcycled models inherit this capability from their Transformer counterparts.In this work, we position long-context preservation as a core objective of upcycling, alongside short-context quality. We introduce our upcycling recipe to convert pretrained Transformer checkpoints into our \underline{\textbf{HY}}brid \underline{\textbf{LO}}ng-context models named \emph{HyLo} without costly pretraining from scratch. Our main contributions are:
\begin{itemize}
    \item \textbf{Long-context-aware model upcycling.} We propose an improved upcycling recipe based on Zebra-Llama~\citep{yang2025zebra} yielding superior long-context performance while having comparable short-context performance (Figure~\ref{fig:fig1},\ref{fig:niah_synthetic}). 
    \item \textbf{Extended long-context training regime.} Prior upcycling studies typically train to around 24K context. We scale staged training from 8K up to 64K tokens and systematically analyze how training sequence length affects long-context generalization.
    \item \textbf{Teacher-guided long-context distillation.} We introduce teacher-guided long-context training with chunk-wise KL supervision, demonstrating significant gains in long-context performance while clarifying the optimization constraints introduced by this distillation design.
    \item \textbf{High throughput inference serving.} We integrate HyLo into \texttt{vLLM}~\citep{kwon2023}, enabling efficient long-context serving with tensor parallelism. HyLo enables serving contexts upto 2M tokens (30$\times$ extension over Llama-3.2-3B) on 8 AMD MI300X GPUs.
 
\end{itemize}

\section{Related Work}
\textbf{Hybrid long-context models trained from scratch.}
Recent work has explored training hybrid architectures from scratch that combine softmax attention with more efficient sequence modeling primitives such as state-space models (SSMs) or linear attention to overcome the quadratic cost of attention. Foundational approaches include S4~\citep{gu2021efficiently} and Mamba~\citep{gu2024mamba}, as well as alternative long-context mechanisms such as RetNet~\citep{sun2023retentive}, Hyena~\citep{poli2023hyena}, and linear attention variants~\citep{katharopoulos2020transformers}. Building on these ideas, recent large-scale hybrids explicitly interleave attention with efficient modules: Jamba~\citep{lieber2024jamba} combines Transformer, Mamba, and MoE layers; Zamba~\citep{glorioso2024zamba} and Samba~\citep{ren2406samba} integrate Mamba with shared or local attention; MiniMax-01 employs Lightning Attention~\citep{qin2024lightning} within a hybrid architecture to enable extreme context scaling~\citep{li2025minimax}; Kimi Linear~\citep{team2025kimi} interleaves Kimi Delta Attention with Multi-Head Latent Attention (MLA)~\citep{liu2024deepseek}; Qwen3-Next~\citep{qwen3next2025} and Qwen3.5~\citep{qwen35_2026} interleave softmax attention with Gated DeltaNet (GDN) ~\citep{yang2024gated} layers. Concurrent work also highlights key design principles: unified positional encoding across attention and SSM components is critical for stability~\citep{wu2025transxssm}, and empirical analyses show that hybrid performance depends heavily on layer allocation, gating, and memory dynamics~\citep{wang2025systematic}, indicating that hybridization requires careful architectural co-design.

\textbf{Post-training upcycling and hybridization.}
An alternative line of work focuses on converting pretrained Transformer models into hybrid long-context models, significantly reducing training cost. Early work MambainLlama~\citep{wang2024mambainllama} shows that pretrained attention layers can initialize SSM blocks, and that retaining a subset of attention layers preserves model quality while enabling length extrapolation. This paradigm is extended by approaches such as Llamba~\citep{bick2025llamba} and X-EcoMLA~\citep{li2025xecomla}, where the latter converts pretrained Transformers into MLA hybrids to improve efficiency and reduce KV-cache overhead. Closely related, Zebra-Llama~\citep{yang2025zebra} combines Mamba2 with MLA and introduces improved initialization, intermediate-layer distillation, and layer selection strategies, achieving near-Transformer performance with limited post-training. Additionally, L2A~\citep{choudhary2026learningattendconditionalmemory} converts softmax attention into sliding window and dynamic full attention hybrid. Subsequent work focuses on identifying which attention components are essential during conversion: methods such as RAD detect redundant attention layers~\citep{hoshino2025rad}, KL-guided approaches optimize hybrid layer allocation~\citep{li2025distilling}, and HALO/HypeNet improve positional adaptation under constrained budgets~\citep{chen2026hybrid}, while retrieval-aware distillation shows that preserving only a small subset of retrieval-critical attention heads is sufficient to maintain long-context reasoning performance~\citep{bick2026retrieval}. Beyond standard language modeling, hybridization has also been explored for reasoning efficiency: the M1 model introduces a hybrid Mamba-based architecture trained with distillation and reinforcement learning, demonstrating that hybrid designs can achieve competitive reasoning performance with improved inference efficiency~\citep{wang2025m1}.

\section{Methodology}
Our goal is to upcycle pretrained Transformer LLMs into long-context hybrid models while preserving short-context quality.
To this end, we propose \ours, an efficient training recipe that extends context length and improves long-range modeling without pretraining from scratch.
Building on MambaInLlama~\citep{wang2024mambainllama} and Zebra-Llama~\citep{yang2025zebra}, which show that careful initialization and staged distillation preserve short-context performance, our method treats long-context preservation as a first-class objective.
Our key contributions are a stronger architecture recipe, staged long-context training, and a broader evaluation across model families and linear block types.

\textbf{Architecture Design} To reduce the quadratic cost of full attention, we use a hybrid architecture that combines \textit{Multi-head Latent Attention (MLA)}~\citep{liu2024deepseek} with linear recurrent blocks, including \textit{Mamba-2} (M2)~\citep{dao2024transformers} and \textit{Gated DeltaNet (GDN)}~\citep{yang2024gated}. The MLA-to-linear ratio defines the quality--efficiency trade-off: more MLA layers increase attention capacity but also increase KV-cache usage, whereas Mamba-2 and GDN add no KV-cache overhead. Unlike prior upcycling studies that focus on one base model and one linear module, we evaluate two Transformer families (Llama and Qwen) and two linear block types (Mamba-2 and GDN), showing that the recipe generalizes across architectures and scales.

\subsection{Initialization}
A key challenge in upcycling is how to initialize replaced hybrid blocks from a pretrained attention-based model.

Following Zebra-Llama~\cite{yang2025zebra}, we first construct a pure MLA model and a pure linear model (Mamba-2 or GDN) by replacing all attention blocks in the base Transformer. Each pure model is then initialized from the original pretrained weights.
Initialization schemes for Mamba-2 and MLA are introduced in MambaInLlama~\citep{wang2024mambainllama} and X-EcoMLA~\citep{li2025x}. Here, we describe our procedure for initializing GDN blocks from Transformer checkpoints.

\textbf{Our GDN Initialization.} In the GDN-based \ours hybrid architecture, each selected decoder layer replaces the standard attention module with a GDN mixer, while preserving the SwiGLU MLP and RMSNorm sublayers from the original Transformer block (see Section~\ref{sec:gdn-init}).

Starting from a pretrained model, each designated GDN layer undergoes in-place module replacement. The MLP weights and RMSNorm parameters are copied verbatim from the corresponding Transformer layer. For attention-to-GDN weight transfer, we address dimension mismatches between the projection weights of the two modules:

\begin{enumerate}[nosep]
    \item \textbf{Grouped-Query Attention (GQA) expansion:} When the teacher uses $H_{\text{kv}} < H_{\text{q}}$ key-value heads (e.g., 8 vs.\ 32 in Llama-3.2-1B), the $K$ and $V$ weight matrices are first expanded by repeating each KV head $H_{\text{q}} / H_{\text{kv}}$ times:
    \begin{equation}
    \tilde{\mathbf{W}}^K = \operatorname{RepeatKV}(\mathbf{W}^K_{\text{teacher}}, g = H_q / H_{\text{kv}}).
    \end{equation}
    \item \textbf{Dimension truncation:} Since GDN's key dimension $d_k < d$ and value dimension $d_v > d$, we transfer the overlapping submatrices:
    \begin{equation}
    \begin{aligned}
    \mathbf{W}^Q_{\text{GDN}}[\,:\!d_k,\, :] &\;\leftarrow\;{\mathbf{W}}^Q_{\text{teacher}}[\,:\!d_k,\, :], \\
    \mathbf{W}^K_{\text{GDN}}[\,:\!d_k,\, :] &\;\leftarrow\; {\mathbf{W}}^K_{\text{teacher}}[\,:\!d_k,\, :], \\
    \mathbf{W}^V_{\text{GDN}}[\,:\!\min(d, d_v),\, :] &\;\leftarrow\; {\mathbf{W}}^V_{\text{teacher}}[\,:\!\min(d, d_v),\, :], \\
    \mathbf{W}^O_{\text{GDN}}[:,\, :\!\min(d, d_v)] &\;\leftarrow\; \mathbf{W}^O_{\text{teacher}}[:,\, :\!\min(d, d_v)].
    \end{aligned}
    \end{equation}
\end{enumerate}
GDN-specific parameters---the gate projection $\mathbf{W}^G$, decay parameters $(\mathbf{A}_{\log}, \Delta_{\text{bias}})$, beta projection $\mathbf{W}_\beta$, and short convolution kernels---remain at their default random initialization.

\subsection{Two-Stage Light Fine-Tuning}
\label{sec:training}
After initialization, we apply two light fine-tuning stages: (i) our enhanced intermediate layer distillation (Enhanced-ILD) and (ii) long context supervised fine-tuning (SFT). In Stage I, pure MLA/Mamba2/GDN models undergo Enhanced-ILD training on only 20\% of the data. We then assemble the final hybrid model from these Stage-I checkpoints and proceed to Stage II.

\textbf{Stage I: Our Enhanced-ILD Training.}
Zebra-Llama~\cite{yang2025zebra} uses ILD to refine initialization by aligning per-layer hidden states. In \ours, we extend this objective by adding an ILD term on token-mixer outputs (i.e., Transformer attention outputs and their corresponding MLA/Mamba2/GDN outputs), which shows a significant improvement to our training.
Therefore, for each layer~$\ell$, we minimize the sum of $L_2$ distances between teacher and student hidden states and token-mixer outputs:
\begin{equation}
\label{eq:ild-loss}
\mathcal{L}_\text{ILD} = \sum_{\ell=1}^{L} \Big[
    \big\| \mathbf{h}_\ell^{(s)} - \mathbf{h}_\ell^{(t)} \big\|_2
    + \big\| \mathbf{a}_\ell^{(s)} - \mathbf{a}_\ell^{(t)} \big\|_2
\Big],
\end{equation}
where $\mathbf{h}_\ell^{(s)}, \mathbf{h}_\ell^{(t)}$ are the student and teacher hidden states after layer~$\ell$, and $\mathbf{a}_\ell^{(s)}, \mathbf{a}_\ell^{(t)}$ are the corresponding attention/token-mixer outputs. This extra ILD term strengthens knowledge transfer from full attention to MLA/Mamba2/GDN blocks; its impact is reported in Table~\ref{tab:ablation_ILD}. We keep the Stage-I context length fixed at 2K.

\textbf{Stage II: Long-Context SFT Training.}
In Stage II, we load the separately distilled MLA/Mamba2/GDN checkpoints from Stage I and assemble them into one hybrid model. Because our focus is long-context extension rather than layer selection, we use either uniform layer selection or baseline-recommended layouts (when available). At this stage, we extend training context length from 2K to 8K and 64K, which is another core contribution.
The assembled hybrid model is then fine-tuned end-to-end with output-level knowledge distillation using KL divergence at extended context lengths:
\begin{equation}
\label{eq:sft-loss}
\mathcal{L}_\text{SFT} =  D_\text{KL}\!\Big(
    \text{softmax}\!\big(\mathbf{z}^{(s)}\big) \;\Big\|\; \text{softmax}\!\big(\mathbf{z}^{(t)}\big)
\Big),
\end{equation}
where $\mathbf{z}^{(s)}$ and $\mathbf{z}^{(t)}$ are the student and teacher logits, respectively.

\begin{table}[t]
\small
\centering
\begin{tabular}{cc}
\hline
Config                                               & Memory (GiB) \\ \hline
No   Teacher                                         & \textbf{OOM} \\
No   Teacher + FusedLinearCE + Act checkpoint        & 137.9        \\
No   Teacher + Act checkpoint                        & 131          \\
No   Teacher + FusedLinearCE + Act checkpoint        & 29.6         \\ \hline
8B   Teacher                                         & \textbf{OOM} \\
8B   Teacher + Fused\_KL                             & \textbf{OOM} \\
8B   Teacher + Act checkpointing                     & \textbf{OOM} \\
8B   Teacher + Fused\_KL\_Hidden                     & 158.8        \\
8B   Teacher + Fused\_KL + Act checkpointing         & 144.8        \\
8B   Teacher + Fused\_KL\_Hidden + Act checkpointing & 54.2         \\ \hline
\end{tabular}
\caption{Training memory for upcycling a Llama-1B model with 4 MLA and 12 Mamba-2 layers at 64K context length, with teacher (KD loss) and without teacher (CE loss).}
\label{tab:train-memory}
\end{table}

\subsection{Memory-Efficient Long-Context Distillation}
Extending knowledge distillation from 2K to 64K context length introduces severe memory pressure.
The dominant bottleneck is the \emph{logit tensor}: for sequence length $T$ and vocabulary size $V$, standard KL divergence requires materializing both student and teacher logits of shape $(T, V)$.
At $T{=}65{,}536$ and $V{=}128{,}256$ (Llama-3), each logit tensor consumes approximately 16\,GB in bfloat16, making naive distillation infeasible even on 80\,GB GPUs.
We address this with progressively stronger memory optimizations (summarized in Table~\ref{tab:train-memory}).
Without a teacher, the 64K setup is still OOM unless we combine activation checkpointing with FusedLinearCE, which reduces memory to 29.6\,GiB. With an 8B teacher, naive KD, Fused\_KL-only, and checkpoint-only settings remain Out of Memory (OOM); combining activation checkpointing with Fused\_KL\_Hidden reduces memory to 54.2\,GiB.
Together, these optimizations enable a $32{\times}$ increase in training context length (2K$\to$64K) with an 8B teacher while maintaining single-epoch training on 8 GPUs.
Further implementation details and a configuration breakdown across context lengths are provided in Appendix~\ref{sec:memory-efficiency}.

\subsection{vLLM Runtime Integration}
 
To enable practical deployment of \ours, we integrate it into \texttt{vLLM}~\citep{kwon2023}. This
requires extending the vLLM inference stack to support architectures that
interleave Mamba/GDN sequence modeling layers with MLA layers, a combination not anticipated by existing serving engines. Our
integration addresses three systems-level challenges: (1)~execution of
heterogeneous layer types (Mamba SSM, GDN linear attention and MLA attention) within a unified serving
engine, where the scheduler must manage both a fixed-size Mamba hidden state and a
variable-size MLA KV cache; (2)~support for MLA-specific KV compression and head
expansion mechanisms, which differ from standard grouped-query attention and
require custom cache allocation logic; and (3)~kernel limitations arising from
model-specific head dimensions (e.g., \ours's compressed latent dimension) that
are not directly supported by existing fused attention implementations such as
FlashAttention, necessitating fallback to PyTorch-based kernels with associated overhead. We implement the required runtime adaptations and evaluate their impact on long-context serving efficiency under paged attention and continuous batching.

\section{Experiments and Results}

\begin{table}[t]
\centering
\scriptsize
\setlength{\tabcolsep}{2pt}
\begin{tabular}{lccc *{7}{c} | *{4}{c} | c}
\toprule
\textbf{Model and Setting} &
\textbf{Teacher} &
\textbf{KV}  &
\multicolumn{8}{c}{\textbf{Common Sense Reasoning $\uparrow$}} &
\multicolumn{4}{c}{\textbf{RULER $\uparrow$}}  &
\textbf{GSM8K $\uparrow$} \\

\cmidrule(lr){4-11}
\cmidrule(lr){12-15}

& & \textbf{cache} &
\textbf{ARC} &
\textbf{ARE} &
\textbf{HS} &
\textbf{OB} &
\textbf{PI} &
\textbf{RA} &
\textbf{WG} &
\textbf{Avg.} &
\textbf{8K} &
\textbf{16K} &
\textbf{32K} &
\textbf{64K} & \\

\hline

\multicolumn{16}{c}{\textbf{Baseline Models}}  \\
\hdashline
MambaInLlama-1B-50\% & 8B & 50\% & 37.7 & 65.5 & 58.2 & 37.6 & 73.2 & 36.5 & 59.3 & \textbf{52.6} &  18.9& 3.0 & 1.0 & 0.0 & 16.2  \\
Llamba 1B & 1B+70B  & 0\% & 37.1 & 65.4 & 61.3 & 36.8 & 73.8 & 37.6 & 60.6 & \textbf{53.2} & 2.9 & 0.0 & 0.0 & 0.0 & 12.5 \\
Zebra-Llama-1B (4MLA-12M2) & 8B & 4\%& 39.1 & 65.4 & 56.9 & 37.0 & 72.3& 34.5 & 57.9 & \textbf{51.8} & 12.3 & 6.8 & 3.7 & 0.1 & 37.2  \\
Zebra-Llama-1B (8MLA-8M2) & 8B & 7.8\%& 38.0 & 66.4 & 58.2 & 38.0 & 72.7 & 36.9 & 61.3 & \textbf{53.1} & 0.5 & 0 & 0.1 & 0 & 43.4  \\
\hline
\multicolumn{16}{c}{\textbf{Training Context Length = $8K$}}  \\
\hdashline
\rowcolor{oursgray}
\textbf{\ours-Llama-4MLA12M2} & 8B & 3.9\% & 38.1 & 65.7 & 57.6 & 37.0 &	72.5 &	35.4 & 58.6 &	\textbf{52.1} & 53.1 & 10.6 & 2.0 & 0.5 & 49.2 \\
\rowcolor{oursgray}
\textbf{\ours-Llama-4MLA12GDN} & 8B & 3.9\%  &38.6 & 66.9 & 59.1 & 37.6 & 72.7 & 36.7 & 60.1 & \textbf{53.1}
 & 55.1 & 11.9 & 2.4 & 0.8 & 51.9  \\
\rowcolor{oursgray}
\textbf{\ours-Llama-8MLA8M2} & 8B & 7.8\%  & 38.8 & 66.7 &	58.3 & 37.0 & 72.3 &	37.4 &	59.7 & \textbf{52.9} 
& 59.0 & 0.3 & 0.1 & 0.0 & 51.0 \\
\rowcolor{oursgray}
\textbf{\ours-Llama-8MLA8GDN} & 8B & 7.8\% & 39.3 & 67.2 & 59.3 & 37.6 & 72.0 & 38.4 & 60.1 & \textbf{53.4} & 60.3 & 0.5 &0.1 &0.1 & 54.6 \\

\hline
\multicolumn{16}{c}{\textbf{Training Context Length = $64K$}}  \\
\hdashline
\rowcolor{oursgray}
\textbf{\ours-Llama-4MLA12M2} & 8B & 3.9\% &35.7 & 63.3 & 55.3 & 34.8 & 71.3 & 34.7 & 56.8 & \textbf{50.3}
& 53.3 & 46.7 & 40.4 & 37.9 & 33.0 \\
\rowcolor{oursgray}
\textbf{\ours-Llama-4MLA12GDN} & 8B & 3.9\% & 36.0 & 63.6 & 57.4 &  38.2 & 70.7 & 35.4 & 57.2 & \textbf{51.2}
 & 52.5 & 48.3 & 44.5 & 40.8 & 37.5\\
\rowcolor{oursgray}
\textbf{\ours-Llama-8MLA8M2} & 8B & 7.8\% & 36.3 & 63.4 & 56.1 & 35.0 & 71.2 & 36.6 & 58.6 & \textbf{51.0}
 & 59.0 & 52.5 & 45.5 & 38.8 & 40.0 \\
\rowcolor{oursgray}
\textbf{\ours-Llama-8MLA8GDN} & 8B & 7.8\% & 36.4 & 64.4 & 57.2 & 37.0 & 72.3 & 37.1 & 58.4 & \textbf{51.8}& 61.5 & 53.7 & 48.1 & 41.6 & 39.4\\

\hline
\end{tabular}
\caption{Comparison of different techniques across backbone models Llama-3.2-1B.}
\label{tab:method_comparison_llama3_1b}
\end{table}

\begin{table}[t]
\centering
\scriptsize
\setlength{\tabcolsep}{2pt}
\begin{tabular}{lccc *{7}{c} | *{4}{c} | c}
\toprule
\textbf{Model and Setting} &
\textbf{Teacher} &
\textbf{KV}  &
\multicolumn{8}{c}{\textbf{Common Sense Reasoning $\uparrow$}} &
\multicolumn{4}{c}{\textbf{RULER $\uparrow$}}  &
\textbf{GSM8K $\uparrow$} \\

\cmidrule(lr){4-11}
\cmidrule(lr){12-15}

& & \textbf{cache} &
\textbf{ARC} &
\textbf{ARE} &
\textbf{HS} &
\textbf{OB} &
\textbf{PI} &
\textbf{RA} &
\textbf{WG} &
\textbf{Avg.} &
\textbf{8K} &
\textbf{16K} &
\textbf{32K} &
\textbf{64K} & \\

\hline
\multicolumn{16}{c}{\textbf{Baseline Models}}  \\
\hdashline
Mamba in Llama-3B-50\% & 70B & 50.0\% & 47.1 & 74.0 & 69.0 & 38.4 & 75.9 & 40.1 & 66.5 & \textbf{58.7} & 37.0 & 1.0 & 0.0 & 0.0 & 56.8\\
Llamba 3B & 3B+70B & 0.0\% & 45.7 & 73.8 & 73.3 & 42.4 & 78.0 & 40.1 & 70.0 & \textbf{60.5} & 3.5 & 0.0 & 0.0 & 0.0 & 47.8 \\
M1 & & 21.4\% & 45.6 & 72.6 & 61.5 & 39.4 & 73.3 & 35.9 & 64.9 & \textbf{56.2} & 63.5 & 43.6 & 30.3 & 17.4 & 62.5\\
Zebra-Llama 3B (6MLA-22M2)  & 8B & 2.0\% & 44.7 & 70.8 &	67.7 & 38.8 & 75.6 &	39.4 & 64.5 & \textbf{57.4}
   & 42.5 & 0.4 & 0.5 & 0.3 & 60.7 \\
Zebra-Llama 3B (14MLA-14M2) & 8B & 4.7\% & 45.7 & 71.8 & 68.6 & 38.6 & 75.7 & 40.9 &	64.4 & \textbf{58.0}
& 35.1 & 13.3 & 6.3 & 4.2 & 66.2 \\
\hline
\multicolumn{16}{c}{\textbf{Training Context Length = $8K$}}  \\
\hdashline
\rowcolor{oursgray}
\textbf{\ours-Llama-6MLA22M2} & 8B & 2.0\% & 45.6 & 72.4 & 67.8 & 38.4 & 76.1 & 39.7 & 66.8 & \textbf{58.1}
 & 65.7 & 39.5 & 25.2 & 11.4 & 66.3 \\
\rowcolor{oursgray}
\textbf{\ours-Llama-6MLA22GDN} & 8B & 2.0\% & 45.4 & 71.9 & 69.3 & 42.4 & 76.6 & 39.7 & 67.6 & \textbf{59.0} & 71.2 & 45.0 & 27.1 & 14.1 & 68.1 \\
\rowcolor{oursgray}
\textbf{\ours-Llama-14MLA14M2}  & 8B & 4.7\% & 46.3 & 73.0 & 68.7 & 40.4 & 75.9 & 40.4 & 67.7 & \textbf{58.9}
 & 75.3 & 49.7 &16.6 & 0.4 & 71.0 \\
\rowcolor{oursgray}
\textbf{\ours-Llama-14MLA14GDN} & 8B & 4.7\% & 47.3 & 72.6 & 69.5 & 40.0 & 76.3 & 41.5 & 67.3 & \textbf{59.2}& 71.1 &45.6 &19.2 &0.2 &68.2  \\

\hline
\multicolumn{16}{c}{\textbf{Training Context Length = $64K$}}  \\
\hdashline
\rowcolor{oursgray}
\textbf{\ours-Llama-6MLA22M2} & 8B & 2.0\% & 43.5 & 69.7 & 66.2 & 38.8 & 75.5 &	38.6 & 64.3 & \textbf{56.7}
 & 65.4 & 56.4 & 49.9 & 42.3 & 51.6  \\
\rowcolor{oursgray}
\textbf{\ours-Llama-6MLA22GDN} & 8B & 2.0\% & 43.7 & 69.5 & 67.9 & 38.6 & 75.9 & 39.8 & 64.9 &  \textbf{57.2} & 68.2 & 62.1 & 55.7 & 46.3 & 56.0 \\
\rowcolor{oursgray}
\textbf{\ours-Llama-14MLA14M2} & 8B & 4.7\% & 44.1 & 71.2 & 67.3 & 39.6 & 75.4 & 40.0 & 64.3 & \textbf{57.4}  & 71.7 & 65.4 & 57.8 & 46.6 & 40.9 \\
\rowcolor{oursgray}
\textbf{\ours-Llama-14MLA14GDN} & 8B & 4.7\% & 45.1 & 72.0 & 68.2 & 39.4 & 76.1 & 40.9 & 63.8 & \textbf{57.9} & 73.2 & 69.7 & 62.9 & 52.0 & 58.9\\
\hline
\end{tabular}
\caption{Comparison of different techniques across backbone models Llama-3.2-3B.}
\label{tab:method_comparison_llama3_3b}
\end{table}

\begin{table}[t]
\centering
\scriptsize
\setlength{\tabcolsep}{2pt}
\begin{tabular}{lccc *{7}{c} | *{4}{c} | c}
\toprule
\textbf{Model and Setting} &
\textbf{Teacher} &
\textbf{KV}  &
\multicolumn{8}{c}{\textbf{Common Sense Reasoning $\uparrow$}} &
\multicolumn{4}{c}{\textbf{RULER $\uparrow$}}  &
\textbf{GSM8K $\uparrow$} \\

\cmidrule(lr){4-11}
\cmidrule(lr){12-15}

& & \textbf{cache} &
\textbf{ARC} &
\textbf{ARE} &
\textbf{HS} &
\textbf{OB} &
\textbf{PI} &
\textbf{RA} &
\textbf{WG} &
\textbf{Avg.} &
\textbf{8K} &
\textbf{16K} &
\textbf{32K} &
\textbf{64K} & \\

\hline
\multicolumn{16}{c}{\textbf{Baseline Models}}  \\
\hdashline
Jet Nemotron-2B & -- & 2.1\%& 42.5& 54.5 & 64.4 & 34.0 & 73.5 & 35.4 & 64.9 & \textbf{52.7} & 71.3 & 60.1  & 43.9 & 14.1 &  19.4 \\
Hype Net (7FA21LA) & 1.7B & 25\% & 41.6 & 67.9 & 57.4 & 36.6 & 72.7 & 32.9 & 63.1 & \textbf{53.2} & 36.4 & 31.3 & 23.8 & 16.4 & 1.1 \\

\hline
\multicolumn{16}{c}{\textbf{Training Context Length = $8K$}}  \\
\hdashline
\rowcolor{oursgray}
\textbf{\ours-Qwen-7MLA21M2} & 8B  & 3.9\% & 44.3 & 71.4 & 60.5 & 39.4 & 73.3 & 36.3 & 61.5 &	\textbf{55.2}
 & 58.7  & 41.1  & 27.5  & 14.6  & 72.3 \\
\rowcolor{oursgray}
\textbf{\ours-Qwen-7MLA21GDN} & 8B & 3.9\% & 45.3 & 72.7 & 61.5 & 39.4 & 73.2 & 36.3 & 64.6 & \textbf{56.1}
&63.5 & 43.6 & 30.3 & 17.4  &76.0 \\
\rowcolor{oursgray}
\textbf{\ours-Qwen-14MLA14M2} & 8B & 7.8\% & 45.9 & 73.2 & 61.2 & 39.2 & 73.8 & 36.5 & 63.1 & \textbf{56.1} & 74.2 & 58.6 & 33.5 & 10.7 & 75.8 \\
\rowcolor{oursgray}
    \textbf{\ours-Qwen-14MLA14GDN} & 8B & 7.8\% & 45.9 & 73.3 & 62.1	& 38.4 &	74.8 &	37.4 &	63.4 & \textbf{56.5}
& 71.1  & 45.6 & 19.2 & 0.2 & 76.1\\
\hline
\multicolumn{16}{c}{\textbf{Training Context Length = $64K$}}  \\
\hdashline
\rowcolor{oursgray}
\textbf{\ours-Qwen-7MLA21M2} & 8B  & 3.9\% & 42.7 & 70.0 &	60.3 & 38.0 & 73.5 & 35.7 & 63.8 & \textbf{54.9}
 & 56.5 & 49.0 & 38.4 & 27.8 & 69.9\\
\rowcolor{oursgray}
\textbf{\ours-Qwen-7MLA21GDN} &  8B & 3.9\% & 44.2 & 71.4 &	61.2 & 37.4 & 73.7 & 36.9 & 63.1 & \textbf{55.4}
&59.8 & 53.8 & 42.5 & 30.5& 73.3 \\
\rowcolor{oursgray}
\textbf{\ours-Qwen-14MLA14M2} & 8B & 7.8\% & 45.4 & 73.3 & 61.1 & 37.8 & 73.7 & 36.6 & 61.8 & \textbf{55.7} & 73.9 & 62.6  & 46.2 & 33.1 & 73.5 \\
\rowcolor{oursgray}
\textbf{\ours-Qwen-14MLA14GDN} & 8B  & 7.8\% & 47.9 & 74.6 & 61.9 & 38.2 & 75.0 & 36.8 &	62.3 &	\textbf{56.7}
 &  66.9 & 53.2 & 41.4 & 31.6 & 73.8 \\
\hline
\end{tabular}
\caption{Comparison of different techniques across backbone models QWEN.}
\label{tab:method_comparison_qwen3}
\end{table}

\subsection{Experimental Setup}

\textbf{Model Configurations.}
We implement our upcycling recipe starting from three base models: Llama-3.2-1B, Llama-3.2-3B, and Qwen3-1.7B. Full model configurations and training hyperparameters are provided in Appendix (Table~\ref{tab:model_configs}).

\textbf{Evaluation Tasks.}
We use the \texttt{lm-eval-harness}~\citep{gao2023} for short context, long context and math reasoning evaluations of our model. For short context common sense reasoning we perform evaluations on language understanding tasks, which includes ARC-Challenge (ARC)~\citep{clark2018think}, ARC-Easy (ARE)~\citep{clark2018think}, HellaSwag (HS)~\citep{zellers2019hellaswag}, OpenBookQA (OB) ~\citep{mihaylov2018can}, PIQA~\citep{bisk2020piqa}, RACE (RA)~\citep{lai2017race}, and WinoGrande (WG) ~\citep{sakaguchi2021winogrande}. For long context evaluations we use all 13 tasks from RULER~\citep{hsieh2024ruler} benchmark. For math reasoning, we include GSM8K~\citep{gsm8k}.  

\textbf{Baselines.} We compare with hybrid model upcycling approaches including MambainLlama\citep{wang2024mambainllama}, Llamba\citep{bick2025llamba}, Zebra-Llama\citep{yang2025zebra}, M1\citep{wang2025m1} and HypeNet\citep{chen2026hybrid}. Among these, HypeNet proposes a hybrid upcycling approach which attempts to maintain long-context performance of the teacher. Additionally, we compare our results with Jet-Nemotron-2B\citep{gu2025jetnemotronefficientlanguagemodel} which is pretrained on 200B tokens from scratch.

\subsection{Main Results}
Tables~\ref{tab:method_comparison_llama3_1b},\ref{tab:method_comparison_llama3_3b},\ref{tab:method_comparison_qwen3} present results across Llama-3.2-1B, Llama-3.2-3B, and Qwen backbone models, comparing HyLo with baseline models on Common Sense Reasoning, GSM8K, and long-context reasoning using RULER. After long-context training, we observe a small drop in short-context performance on Common Sense Reasoning benchmarks for some models, which is expected when adapting models to longer context lengths. However, the performance degradation is relatively small across all backbone models and training settings. GSM8K performance remains competitive across most configurations.

In contrast, we observe substantial improvements on long-context reasoning tasks. HyLo models significantly outperform baseline models on the RULER benchmark, particularly at longer evaluation context lengths such as 32K and 64K tokens. 
Our models maintain much stronger performance as context length increases, suggesting that the proposed training recipe improves the model’s ability to effectively generalize over long context.

We also compare models trained with different training context lengths (8K and 64K). Models trained with longer training contexts generally achieve better performance on longer evaluation contexts while showing only modest decreases in short-context reasoning performance (also observed in \citep{gao2025train}). Overall, the results demonstrate that our long-context training recipe effectively improves long-context reasoning while maintaining strong short-context and mathematical reasoning performance across multiple backbone models.

\begin{figure}[t]
    \centering
    \includegraphics[width=0.8\linewidth]{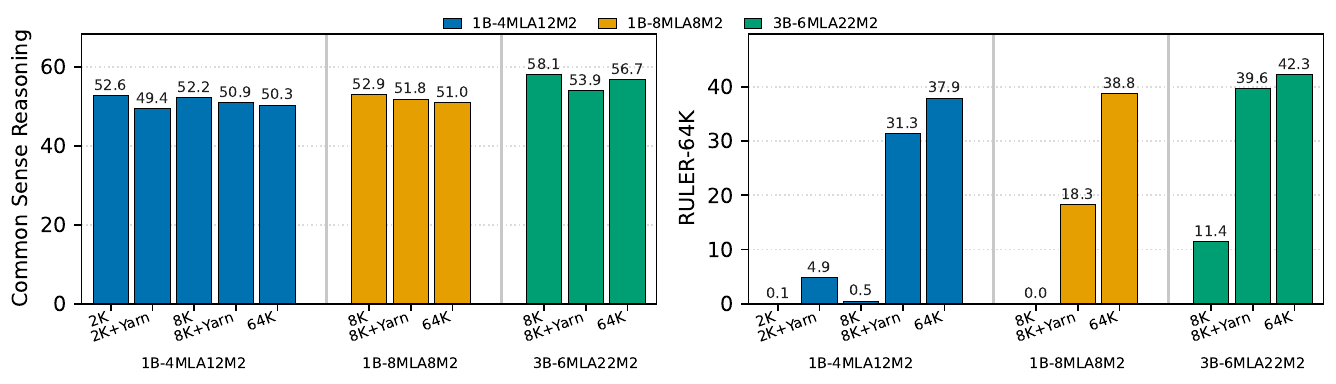}
    \caption{Impact of training sequence length and position interpolation using Yarn. Applying Yarn extension improves long context performance with a slight degradation in short context commonsense reasoning abilities. Furthermore, training at longer context preserves the long context abilities to a greater extent.}
    \label{fig:seqlen_yarn}
\end{figure}
\begin{figure}[t]
    \centering
    \includegraphics[width=0.75\linewidth]{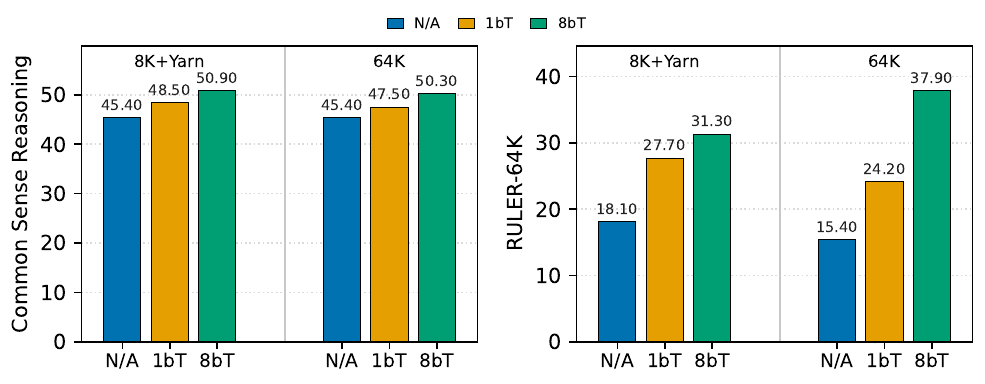}
    \caption{Impact of size of teacher at long context knowledge distillation. Larger teacher improves both short-context common sense reasoning tasks as well as long context ability.}
    \label{fig:kd_teacher_size}
\end{figure}
\subsection{Ablation Studies}

\paragraph{Comparison with position interpolation.}
 
To reduce the computational cost of model upcycling, we train models with shorter context lengths and then apply zero-shot context length extension. Specifically, we train models at different sequence lengths while keeping the total training token budget constant, and then apply YaRN position interpolation~\citep{peng2023yarn} to the RoPE embeddings in MLA layers to extend the context length. Mamba layers, which do not use positional embeddings, remain unchanged. In Figure~\ref{fig:seqlen_yarn}, we evaluate performance on both short- and long-context tasks. Applying YaRN slightly reduces short-context accuracy but significantly improves long-context performance. For example, the 1B-4MLA-12M2 model trained with an 8K context achieves 50.7\% average accuracy on short-context tasks but only 0.5\% on RULER at 64K. After YaRN scaling, short-context accuracy decreases slightly to 49.0\%, while long-context performance improves to 31.3\% at 64K. Importantly, we observe that training directly with a 64K context length yields the best long-context performance while maintaining comparable short-context accuracy. Similar trends are observed for the 1B-8MLA-8M2 and 3B-6MLA-22M2 models. These results demonstrate that long-context training is effective for hybrid model upcycling.

\textbf{Impact of knowledge distillation.} 
 
We analyze the effectiveness of knowledge distillation (KD) in our long-context training recipe. While prior work (e.g., MambaInLlama~\citep{wang2024mambainllama} and Zebra Llama~\citep{yang2025zebra}) showed KD improves short-context performance, we study its impact on long-context learning. As shown in Figure~\ref{fig:kd_teacher_size}, KD has a substantially larger effect on long-context performance than on short-context tasks. For the 1B-4MLA-12M2 model trained at 64K context length, using an 8B teacher improves short-context reasoning accuracy by 6\%, while RULER accuracy at 64K improves by 22\%. Larger teacher models consistently yield greater gains. When training at 8K context length and extending context using YaRN, KD still improves RULER-64K accuracy by 14\%, showing KD remains effective even when long-context ability is obtained via post-training context extension. Overall, combining KD with long-context training significantly improves performance under the same training token budget.

\begin{table}[t]
\centering
\scriptsize
\setlength{\tabcolsep}{2pt}
\begin{tabular}{lc *{7}{c} | *{5}{c}  }
\toprule
\textbf{Model and Setting} &
\multicolumn{8}{c}{\textbf{Common Sense Reasoning $\uparrow$}} &
\multicolumn{5}{c}{\textbf{RULER $\uparrow$}}  \\

\cmidrule(lr){2-9}
\cmidrule(lr){10-14}

& 
\textbf{ARC} &
\textbf{ARE} &
\textbf{HS} &
\textbf{OB} &
\textbf{PI} &
\textbf{RA} &
\textbf{WG} &
\textbf{Avg.} &
\textbf{4K} &
\textbf{8K} &
\textbf{16K} &
\textbf{32K} &
\textbf{64K}  \\

\midrule
1B-4MLA12M2 & 36.6 & 64.3 & 55.5 & 35.6 & 71.4 & 35.5 & 57.5 & 49.1 & 50.6 & 44.1 & 41.6 & 38.6 & 31.3  \\
1B-4MLA12M2 w/ attn. gating &37.0 & 63.9 & 54.7 & 34.4 & 70.3  & 35.6 & 57.2 & 48.6 & 52.4 & 44.3 & 41.9 & 38.5 & 29.3 \\
1B-4MLA12M2 w/ NoPE & 38.5 & 66.5 & 57.0 &  36.0 & 71.8 & 35.0 & 56.8 & 49.9 & 59.2 & 51.1  & 4.8 & 1.4 & 0.0 \\

\bottomrule
\end{tabular}
\caption{Ablation on architectural design choices incorporated in our upcycled hybrid models. (a) Removes utilizes No Positional Embeddings (NoPE) in MLA layers,while (b) adds learnable gating after attention output in MLA layers. While both NoPE and attention gating have been shown to improve long context generalization, same trends do not hold for our upcyled hybrid models.}
\label{tab:ablation_arch}
\end{table}

\textbf{Ablation on architectural design choices.}

Recently, several architectural modifications have been proposed to improve long-context performance. No Position Embedding (NoPE)~\citep{yang2025rope2nope} removes positional information from full attention layers and improves extrapolation beyond the training length. DRoPE~\citep{2026drope} further shows that pretraining with RoPE followed by finetuning with NoPE yields strong performance. Another orthogonal approach, Gated Attention~\citep{qiu2025gatedattentionlargelanguage}, applies a multiplicative learnable sigmoid gate to attention outputs and has also been shown to improve long-context extrapolation. Motivated by these works, we evaluate NoPE and Gated Attention in our long-context hybrid model upcycling framework. We train models at 8K context length and extend context using YaRN. However, neither method improves performance in our setting. Applying NoPE to MLA layers yields no long-context generalization gains. Gated Attention provides small improvements at 4K, 8K, and 16K, but the gains diminish at longer contexts, and performance at 64K is 1\% lower than our baseline 1B-4MLA-12M2 model trained at 8K without these modifications. These results suggest that while NoPE and Gated Attention are beneficial when included during pretraining, they do not provide improvements in the hybrid upcycling setting.

\begin{table}[t]
\centering
\scriptsize
\setlength{\tabcolsep}{2pt}
\begin{tabular}{lc *{7}{c} | *{1}{c}  }
\toprule
\textbf{Model and Setting} &
\multicolumn{8}{c}{\textbf{Common Sense Reasoning $\uparrow$}} &
\multicolumn{1}{c}{\textbf{GSM8K $\uparrow$}}  \\

\cmidrule(lr){2-9}

& 
\textbf{ARC} &
\textbf{ARE} &
\textbf{HS} &
\textbf{OB} &
\textbf{PI} &
\textbf{RA} &
\textbf{WG} &
\textbf{Avg.} &
  \\

\midrule

1B-4MLA12M2 & 39.1 & 65.4 & 56.9 & 37.0 & 72.3 & 34.5 & 57.9 & \textbf{51.8} & \textbf{37.2}  \\
\rowcolor{oursgray}
1B-4MLA12M2 \textbf{+ Our Enhanced-ILD} &  38.7 & 66.7	& 57.9 & 37.8 & 72.7 & 36.4 & 59.0 & \textbf{52.8}
  & \textbf{43.5}  \\
1B-8MLA8M2 & 38.0 & 66.4 & 58.2 & 38.0 & 72.7 & 36.9 & 61.3 & \textbf{53.1} & \textbf{43.4}  \\
\rowcolor{oursgray}
1B-8MLA8M2 \textbf{+ Our Enhanced-ILD} & 37.5 & 66.9 &	58.6 &	38.0 &	73.6 & 37.6 &	61.6 & \textbf{53.4} & \textbf{48.8}  \\
8B-8MLA24M2 & 52.1 & 77.1 & 74.3 &	41.8 & 78.8 & 40.8 &	69.7 & \textbf{62.1} & \textbf{66.3}  \\
\rowcolor{oursgray}
8B-8MLA24M2\textbf{+ Our Enhanced-ILD} & 52.1 & 76.9 & 74.5 &	42.4 & 79.0 &	42.5 &	69.1 & \textbf{62.3} & \textbf{72.4} \\
\bottomrule
\end{tabular}
\caption{Ablation of the impact of Enhanced-ILD loss.}
\label{tab:ablation_ILD}
\end{table}
 
\textbf{Impact of our Enhanced-ILD loss.}
In this ablation, we evaluate the effect of our Enhanced Intermediate-Layer Distillation (Enhanced-ILD) loss, which aligns token-mixing representations between the Transformer teacher and the corresponding hybrid student blocks (MLA/M2/GDN). Table~\ref{tab:ablation_ILD} shows results for models trained with the regular ILD loss introduced in Zebra-Llama~\citep{yang2025zebra} versus our Enhanced-ILD loss. We observe consistent improvements across model scales and hybrid compositions. On commonsense reasoning, Enhanced-ILD yields stable gains in average score: from 51.8 to 52.8 (+1.0) for 1B (4MLA-12M2), from 53.1 to 53.4 (+0.3) for 1B (8MLA-8M2), and from 62.1 to 62.3 (+0.2) for 8B (8MLA-24M2). More importantly, Enhanced-ILD provides a much larger and more consistent boost on GSM8K: 37.2 to 43.5 (+6.3), 43.4 to 48.8 (+5.4), and 66.3 to 72.4 (+6.1), respectively. These results indicate that Enhanced-ILD is especially effective for strengthening mathematical reasoning while preserving, or slightly improving, broad commonsense performance.

\begin{figure}[t]
    \centering
    \includegraphics[width=0.7\linewidth]{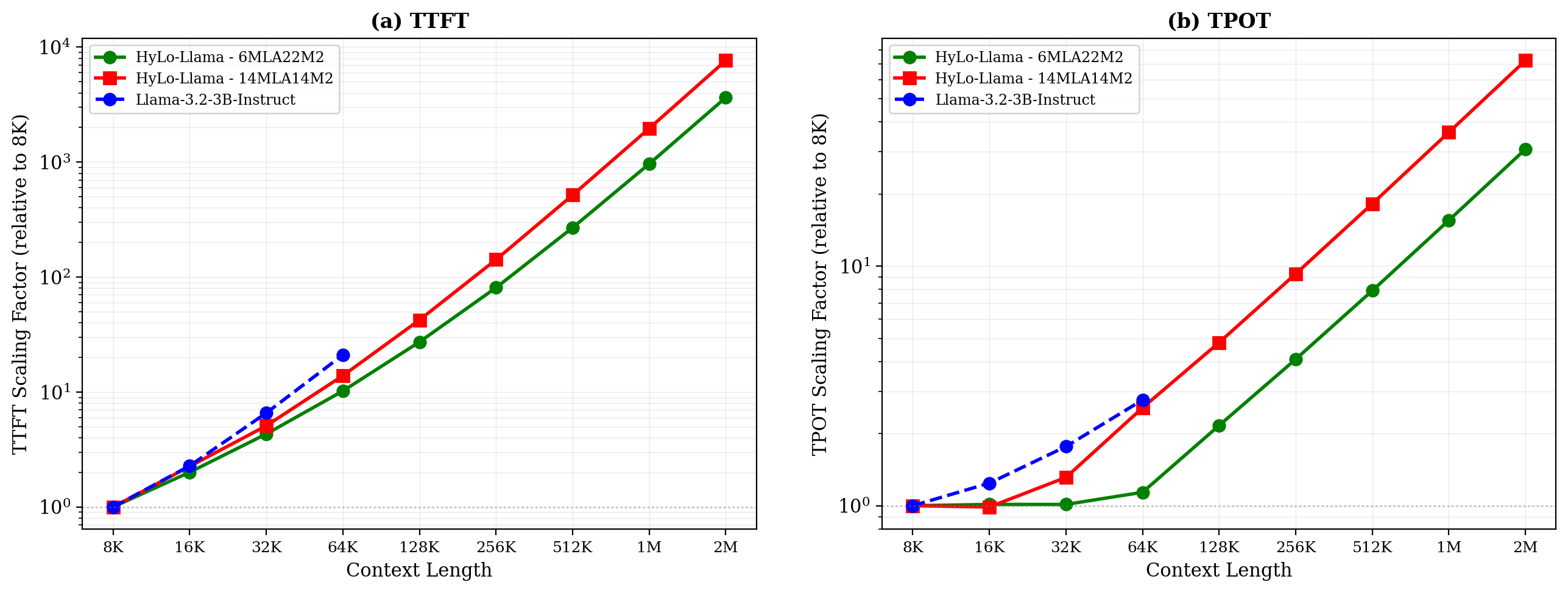}
    \caption{TTFT and TPOT comparison for 3B models with backbone model Llama-3.2-3B on vLLM.}
    \label{fig:inference_results}
\end{figure}
 
\subsection{Inference Latency Evaluation}
 
All experiments reported use vLLM with TP=8, batch size=1,  on a single node with 8 AMD Instinct MI300X GPUs. Each model is tested on a context-length sweep from
8K to 2M tokens, measuring prefill and  decode latency.

\textbf{Prefill latency.}
Figure~\ref{fig:inference_results} reports TTFT across context lengths. At 8K--64K,
all three models show comparable prefill latency. Beyond 64K, Llama 3B runs OOM, as its 28 attention layers each maintain a full KV cache whose combined footprint exceeds GPU memory. Both HyLo
variants complete the full sweep up to 2M context length: while \ours-Llama-6MLA22M2 is about 2.2x faster than  \ours-Llama-14MLA14M2 at 2M, directly reflecting
the O($n^2$) cost of 14 vs.\ 6 MLA layers. 

\textbf{Decode latency.}
Figure~\ref{fig:inference_results} shows per-token decode latency. At short contexts
(8K--32K), Llama 3B achieves lower TPOT. However, its TPOT grows linearly with context as KV cache
access scales across all 28 layers, and the model OOMs at 128K. \ours-Llama-6MLA22M2 maintains a flat TPOT from 8K through 64K, as the Mamba layers use a fixed-size hidden state rather than an expanding cache. Beyond 64K, TPOT rises
sub-linearly as the MLA layers' KV cache grows and at 2M, \ours-Llama-6MLA22M2 reaches about 2x faster throughput compared with \ours-Llama-14MLA14M2.

\section{Conclusion}
 
In this paper, we present \emph{HyLo}, a series of hybrid LLMs upcycled from pretrained Transformer checkpoints, with explicit emphasis on preserving long-context capability. We introduced long-context aware upcycling strategy that combines MLA-based transformer attention blocks with linear blocks instantiated with both Mamba2 and GDN, staged context-length expansion and teacher guided distillation. Across 1B- and 3B-scale settings, including both Llama- and Qwen-based backbones, our results indicate that HyLo achieves superior long-context generalization while maintaining competitive short-context quality compared to related hybrid model upcycling baselines. Additionally, Beyond quality, HyLo is deployment-oriented: our models reduce KV-cache memory by more than $90\%$ and, with our integrated \texttt{vLLM} runtime, support up to 2M-token prefill and decoding. 
As future work, we plan to further close the remaining gap at long context lengths, improve distillation efficiency, and extend this framework to broader downstream settings where robust long-context reasoning is essential.

\bibliography{colm2026_conference}
\bibliographystyle{colm2026_conference}

\clearpage
\appendix
\section{Appendix}
\subsection{More Experimental Details}
\subsubsection{Details of Model Configurations}
\label{sec:model-configs}

We implement our upcycling recipe starting from three different base models: Llama-3.2-1B, Llama-3.2-3B, and Qwen3-1.7B. Notably, Qwen3 utilizes normalization after the Query and Key projection layers, which we maintain when performing attention layer conversion. Table~\ref{tab:model_configs} summarizes the model configurations and hyperparameters used in our experiments.

\subsubsection{Enhanced-ILD hyperparameters.}
MLA/Mamba2/GDN Enhanced-ILD runs share the same configuration: 1~epoch on 20\% of the training SFT dataset~\citep{wang2024mambainllama}, with context length 2048, learning rate $2{\times}10^{-4}$ with cosine decay, warmup ratio 0.01, and bfloat16 mixed precision. Training is distributed across 8 AMD MI300X GPUs using FSDP with full sharding.

\subsubsection{SFT hyperparameters}
We train for 1~epoch at context length 8K/64K, learning rate as reported in Table~\ref{tab:model_configs} with cosine schedule, warmup ratio 0.01, and the full dataset ($\text{data\_ratio}{=}1.0$). YaRN-based position scaling extends the effective context from the original 2048 to 8192 tokens (scaling factor~4.0) or from the original 2048 to 65536 (scaling factor~32.0). Training uses FSDP across 8 AMD MI300X GPUs.

\begin{table}[h]
\centering
\scriptsize
\scalebox{0.95}{
\begin{tabular}{llccccccc}
\toprule
Our Models & Base Model & MLA Layer Indices & \makecell[c]{\# Act.\\ Params.} & Head & Layer & Hidden & lr & \makecell[c]{batch size\\(8k/64k)} \\
\midrule
\ours-Llama-4MLA12M2 & Llama-3.2-1B & [1,5,10,14]& 1.5B & 32 & 16 & 2048  & $6.0 \times 10^{-5}$ & 32/8 \\
HyLo-Llama-4MLA12GDN & Llama-3.2-1B & [1,5,10,14] & 1.7B & 32 & 16 & 2048  & $6.0 \times 10^{-5}$ & 32/8 \\
HyLo-Llama-8MLA8M2 &   Llama-3.2-1B & [0,2,4,6,8,10,12,14] & 1.5B & 32 & 16 & 2048  & $6.0 \times 10^{-5}$ & 32/8 \\
HyLo-Llama-8MLA8GDN &   Llama-3.2-1B & [0,2,4,6,8,10,12,14] & 1.6B & 32 & 16 & 2048  & $6.0 \times 10^{-5}$ & 32/8 \\
HyLo-Llama-6MLA22M2 &   Llama-3.2-3B &[0,5,10,16,21,26]& 3.8B & 24 & 28 & 3072  & $4.0 \times 10^{-5}$ & 16/8 \\
HyLo-Llama-6MLA22GDN &   Llama-3.2-3B &[0,5,10,16,21,26]& 4.3B & 24 & 28 & 3072  & $4.0 \times 10^{-5}$ & 16/8 \\
HyLo-Llama-14MLA14M2 &   Llama-3.2-3B &\makecell[c]{[0,2,4,6,8,10,12,14,\\16,18,20,22,24,26]} & 3.7B & 24 & 28 & 3072  & $4.0 \times 10^{-5}$ & 16/8 \\
HyLo-Llama-14MLA14GDN &   Llama-3.2-3B &\makecell[c]{[0,2,4,6,8,10,12,14,\\16,18,20,22,24,26]}  & 4.0B & 24 & 28 & 3072  & $4.0 \times 10^{-5}$ & 16/8 \\
HyLo-Qwen-7MLA21M2 &   Qwen3-1.7B & [1,5,9,13,17,21,25]& 2.1B & 16 & 28 & 2048  & $6.0 \times 10^{-5}$ & 32/8 \\
HyLo-Qwen-7MLA21GDN &   Qwen3-1.7B & [1,5,9,13,17,21,25] & 2.3B & 16 & 28 & 2048  & $6.0 \times 10^{-5}$ & 16/8 \\
HyLo-Qwen-14MLA14M2 &   Qwen3-1.7B & \makecell[c]{[0,2,4,6,8,10,12,14,\\16,18,20,22,24,26]} & 2.1B & 16 & 28 & 2048  & $6.0 \times 10^{-5}$ & 32/8 \\
HyLo-Qwen-14MLA14GDN &   Qwen3-1.7B & \makecell[c]{[0,2,4,6,8,10,12,14,\\16,18,20,22,24,26]} & 2.2B & 16 & 28 & 2048  & $6.0 \times 10^{-5}$ & 16/8 \\
\bottomrule
\end{tabular}
}
\caption{Model configurations and hyperparameters for our experiments.}
\label{tab:model_configs}
\end{table}

\subsection{MLA Layer Architecture and SVD-Based Initialization}
\label{sec:mla-init}

The Multi-head Latent Attention (MLA) layers in \ours follow the DeepSeek-V3 design~\citep{liu2024deepseek}, which compresses the key-value cache through low-rank latent projections. We initialize MLA layers from pretrained Transformer attention weights using SVD-based decomposition following the methodology outlined in X-EcoMLA~\citep{li2025x}.

\subsubsection{MLA Architecture}

Given input $\mathbf{x}_t \in \mathbb{R}^d$, the MLA module computes queries, keys, and values through two low-rank bottlenecks:

\paragraph{Query path.}
\begin{equation}
\label{eq:mla-query}
\begin{aligned}
\mathbf{c}_t^Q &= \mathbf{W}^{QA} \mathbf{x}_t \in \mathbb{R}^{r_q}, \\
\mathbf{q}_{t}^{\text{nope}} &= \mathbf{W}^{QB}\, \text{Norm}(\mathbf{c}_t^Q) \in \mathbb{R}^{H \times d_{qk}^{\text{nope}} },\\
\mathbf{q}_{t}^{\text{rope}} &= \mathbf{W}^{QR}\, \text{Norm}(\mathbf{c}_t^Q) \in \mathbb{R}^{H \times d_{qk}^{\text{rope}}},
\end{aligned}
\end{equation}
where $r_q$ is the query latent rank, and $\mathbf{q}_t^{\text{rope}}$ receives RoPE.

\paragraph{Key-value path.}
\begin{equation}
\label{eq:mla-kv}
\begin{aligned}
\mathbf{c}_t^{KV} &= \mathbf{W}^{KVA} \mathbf{x}_t \in \mathbb{R}^{r_{kv}}, \\
\mathbf{k}_t^{\text{rope}} &= \mathbf{W}^{KR} \mathbf{x}_t \in \mathbb{R}^{ d_{qk}^{\text{rope}}}, \\
\mathbf{k}_{t}^{\text{nope}} &= \mathbf{W}^{KB}\, \text{Norm}(\mathbf{c}_t^{KV}) \in \mathbb{R}^{H_{kv} \times d_{qk}^{\text{nope}} }, \\
\mathbf{v}_{t} &= \mathbf{W}^{VB}\, \text{Norm}(\mathbf{c}_t^{KV}) \in \mathbb{R}^{H_{kv} \times d_v},
\end{aligned}
\end{equation}
where $r_{kv}$ is the joint key-value latent rank. The KV cache only stores the compressed latent $\mathbf{c}_t^{KV} \in \mathbb{R}^{r_{kv}}$ and the rope key $\mathbf{k}_t^{\text{rope}} \in \mathbb{R}^{d_{qk}^{\text{rope}}}$, reducing the per-token cache from $2 H_{kv} d_h$ to $r_{kv} + d_{qk}^{\text{rope}}$.

\paragraph{Attention and output.}
The full query and key are assembled as $\mathbf{q}_t = [\mathbf{q}_t^{\text{nope}};\; \text{RoPE}(\mathbf{q}_t^{\text{rope}})]$ and $\mathbf{k}_t = [\mathbf{k}_t^{\text{nope}};\; \text{RoPE}(\mathbf{k}_t^{\text{rope}})]$, then standard scaled dot-product attention is applied:
\begin{equation}
\label{eq:mla-attn}
\mathbf{o}_t = \mathbf{W}^O \operatorname{Attn}(\mathbf{q}_t, \mathbf{k}_t, \mathbf{v}_t).
\end{equation}

\subsubsection{SVD-Based Initialization from Transformer Weights}
\label{sec:mla-svd-init}

To initialize the low-rank MLA projections from a pretrained Transformer model, we decompose the teacher's full-rank attention weights via truncated SVD.

\paragraph{Query initialization.}
Let $\mathbf{W}^Q \in \mathbb{R}^{(H \cdot d_h) \times d}$ be the teacher's query projection. We compute its SVD:
\begin{equation}
\label{eq:svd-q}
\mathbf{W}^Q = \mathbf{U}_Q \, \boldsymbol{\Sigma}_Q \, \mathbf{V}_Q^\top,
\end{equation}
and initialize the MLA down/up projections as:
\begin{equation}
\label{eq:mla-q-init}
\begin{aligned}
\mathbf{W}^{QA} &\leftarrow \boldsymbol{\Sigma}_Q[:r_q] \, \mathbf{V}_Q[:r_q, :]^\top \in \mathbb{R}^{r_q \times d}, \\
\mathbf{W}^{QB} &\leftarrow \operatorname{Select}(\mathbf{U}_Q[:, :r_q],\; d_{qk}^{\text{nope}},\; d_{qk}^{\text{rope}}) \in \mathbb{R}^{(H \cdot d_{qk}) \times r_q},
\end{aligned}
\end{equation}
where $\operatorname{Select}(\cdot)$ reshapes $\mathbf{U}_Q$ into per-head blocks and retains only the first $d_{qk}^{\text{nope}}$ and last $d_{qk}^{\text{rope}}$ dimensions from each head's $d_h$-dimensional slice, discarding the middle dimensions that are not used in MLA.

\paragraph{Joint key-value initialization.}
Key-value initialization is complicated by MLA's decoupled RoPE design. When the teacher uses GQA ($H_{kv} < H$), we first expand $\mathbf{W}^{K}$ and $\mathbf{W}^{V}$ to $H$ heads by replicating each KV group $H/H_{kv}$ times, then apply truncated SVD to the concatenated matrix:
\begin{equation}
    [\mathbf{W}^{K}, \mathbf{W}^{V}] = \mathbf{U}_{KV}\boldsymbol{\Sigma}_{KV}\mathbf{V}_{KV}^{\top}.
\end{equation}
We set
\begin{equation}
    \mathbf{W}^{KVA}=\mathbf{U}_{KV}[:, :r_{kv}], \quad
    \mathbf{W}^{KVB}=\boldsymbol{\Sigma}_{KV}[:r_{kv}]\,\mathbf{V}_{KV}[:r_{kv}, :]^{\top}.
\end{equation}
With $d_v=d_h$, we split $\mathbf{W}^{KVB}$ into key and value parts (same column order as $[\mathbf{W}^K,\mathbf{W}^V]$):
\begin{equation}
\begin{aligned}
    \mathbf{W}^{VB} &= \mathbf{W}^{KVB}[:,\, H_{kv}d_h:], \\
    \bar{\mathbf{W}}^{KB} &= \operatorname{reshape}(\mathbf{W}^{KVB}[:, :H_{kv}d_h]) \in \mathbb{R}^{r_{kv}\times H_{kv}\times d_h}, \\
    \mathbf{W}^{KB} &= \operatorname{reshape}(\bar{\mathbf{W}}^{KB}[:, :, :d_{qk}]).
\end{aligned}
\end{equation}
Finally, because all MLA heads share the same RoPE key embedding, we initialize $\mathbf{W}^{KR}$ from the head-averaged key projection $\mathbf{W}^{K}_{\text{avg}}$:
\begin{equation}
    \mathbf{W}^{KR}=\mathbf{W}^{K}_{\text{avg}}[:, -d_r:].
\end{equation}

\paragraph{Output projection.}
The output projection is truncated from the teacher:
\begin{equation}
\mathbf{W}^O \leftarrow \mathbf{W}^O[:, :H \cdot d_v] \in \mathbb{R}^{d \times (H \cdot d_v)}.
\end{equation}

\paragraph{MLP and layer norms.} All MLP weights and RMSNorm parameters are copied directly from the teacher.

\begin{figure}[t]
    \centering
    \includegraphics[width=0.9\linewidth]{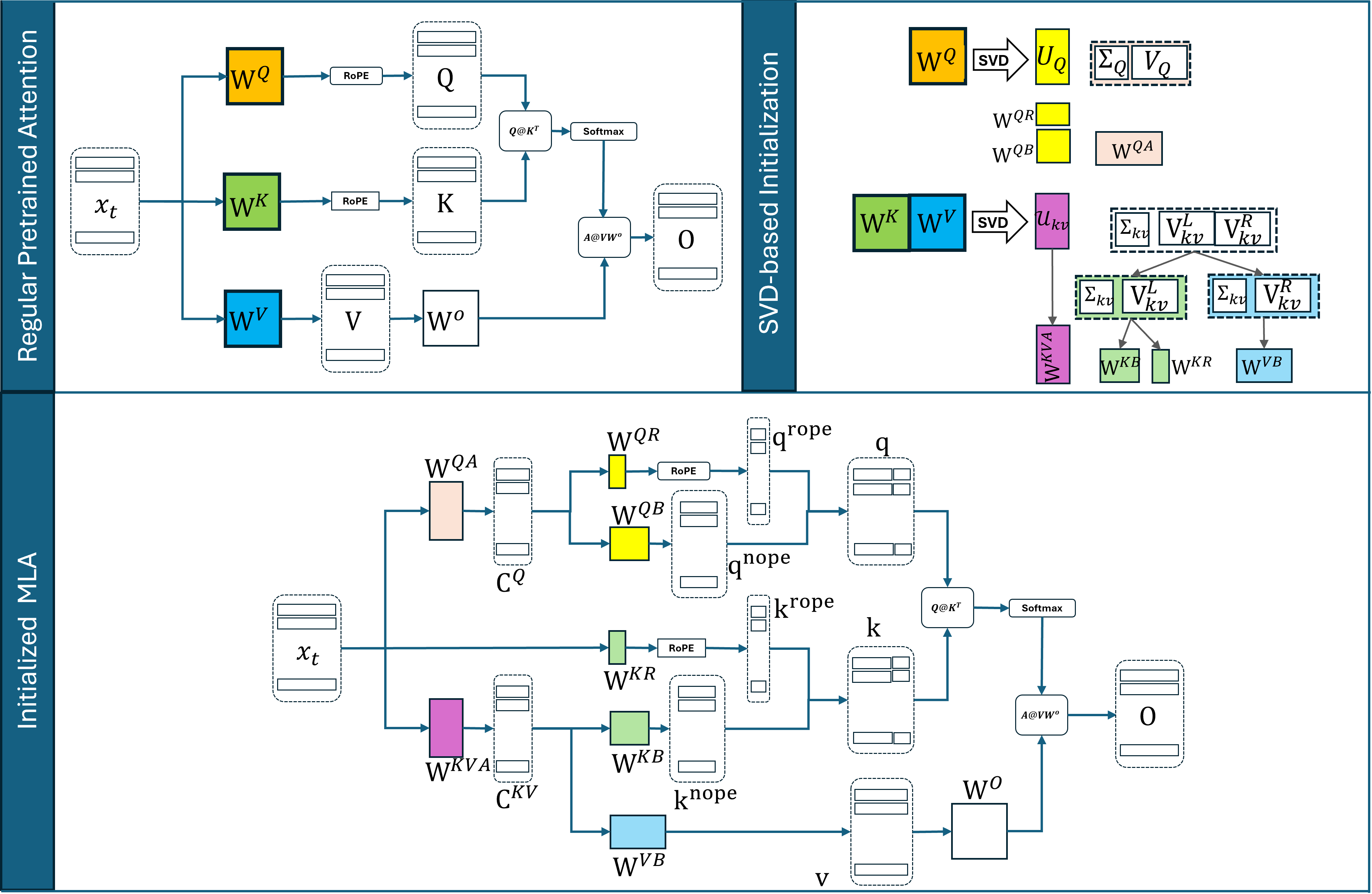}
    \caption{Overview of MLA initialization from a pretrained Transformer attention block.}
    \label{fig:mla_init}
\end{figure}

\subsection{GDN Layer Architecture}
\label{sec:gdn-init}

In the GDN-based \ours hybrid architecture, each non-attention decoder layer replaces the standard attention module with a \textbf{Gated DeltaNet (GDN)}~\citep{yang2024gated} mixer while preserving the SwiGLU MLP and RMSNorm sub-layers from the original transformer block. Concretely, a GDN decoder layer consists of:
\begin{equation}
\label{eq:gdn-layer}
\begin{aligned}
\mathbf{h}' &= \mathbf{h} + \operatorname{GDN}\!\bigl(\operatorname{RMSNorm}(\mathbf{h})\bigr), \\
\mathbf{h}'' &= \mathbf{h}' + \operatorname{MLP}\!\bigl(\operatorname{RMSNorm}(\mathbf{h}')\bigr),
\end{aligned}
\end{equation}
where $\operatorname{GDN}(\cdot)$ is the Gated DeltaNet module from FLA~\citep{yang2024fla}.

\paragraph{GDN mixer parameterization.}
With the gating mechanism enabled (\texttt{use\_gate=True}), the GDN mixer allocates parameters as follows. Let $d$ denote the model hidden size. The key dimension is $d_k = \lfloor 0.75 \cdot d \rfloor$, distributed over $H$ heads each of dimension $d_h = d_k / H$. The value dimension is $d_v = \alpha \cdot d_k$ with expansion ratio $\alpha = 2$. The projections are:
\begin{itemize}[nosep]
    \item $\mathbf{W}^Q, \mathbf{W}^K \in \mathbb{R}^{d_k \times d}$ \quad ($0.75 \, d^2$ parameters each),
    \item $\mathbf{W}^V, \mathbf{W}^G, \mathbf{W}^O \in \mathbb{R}^{d_v \times d}$ \quad ($1.5 \, d^2$ parameters each),
    \item $\mathbf{W}_\alpha, \mathbf{W}_\beta \in \mathbb{R}^{H \times d}$ \quad (decay and beta projections),
    \item $\mathbf{A}_{\log} \in \mathbb{R}^{H}$, $\Delta_{\text{bias}} \in \mathbb{R}^{H}$ \quad (learnable decay and timestep biases),
\end{itemize}
yielding approximately $6 \, d^2$ parameters per layer.
Each of $Q$, $K$, and $V$ is processed through a short 1-D convolution (kernel size 4) with SiLU activation before the recurrence.

\paragraph{Gated Delta Rule Recurrence.}
The GDN mixer maintains a per-head state matrix $\mathbf{S}_t \in \mathbb{R}^{d_k \times d_v}$ that is updated at every timestep $t$ via the \emph{gated delta rule}:
\begin{align}
    \tilde{\mathbf{S}}_t &= e^{g_t} \cdot \mathbf{S}_{t-1},
        \label{eq:gdn-decay} \\[4pt]
    \mathbf{v}'_t &= \mathbf{v}_t - \tilde{\mathbf{S}}_t^{\!\top} \mathbf{k}_t,
        \label{eq:gdn-delta} \\[4pt]
    \mathbf{S}_t &= \tilde{\mathbf{S}}_t + \mathbf{k}_t \bigl(\beta_t \cdot \mathbf{v}'_t\bigr)^{\!\top},
        \label{eq:gdn-write} \\[4pt]
    \mathbf{o}_t &= \frac{1}{\sqrt{d_k}}\, \mathbf{S}_t^{\!\top} \mathbf{q}_t,\\
        \label{eq:gdn-read}
    \mathbf{o}_t &= \operatorname{RMSNorm}\bigl(\mathbf{q}_t \, \mathbf{S}_t,\; \mathbf{W}_G \mathbf{x}_t\bigr), \\
    \mathbf{y}_t &= \mathbf{W}_O \, \mathbf{o}_t,
\end{align}
where $g_t \in (-\infty, 0)$ is the per-head forget gate (Eq.~\ref{eq:gdn-decay}), $\beta_t \in (0, 1)$ is the per-head write strength, $\mathbf{q}_t, \mathbf{k}_t \in \mathbb{R}^{d_k}$ are queries and keys, $\mathbf{v}_t \in \mathbb{R}^{d_v}$ are values, and $d_k = \lfloor 0.75 \, d \rfloor$, $d_v = 2\,d_k$.

Eq.~\eqref{eq:gdn-decay} applies an exponential decay to the state, controlled by $g_t = -\exp(\mathbf{A}_{\log}) \cdot \mathrm{softplus}(\mathbf{W}_\alpha \mathbf{x}_t + \Delta_{\mathrm{bias}})$.
Eq.~\eqref{eq:gdn-delta} is the \emph{delta correction}: it retrieves the value currently associated with key~$\mathbf{k}_t$ and subtracts it from the new value~$\mathbf{v}_t$, preventing superposition interference.
Eq.~\eqref{eq:gdn-write} writes the corrected value back into the state, scaled by $\beta_t = \sigma(\mathbf{W}_\beta \mathbf{x}_t)$.
Eq.~\eqref{eq:gdn-read} reads the output by querying the updated state with~$\mathbf{q}_t$.

During training, the sequential recurrence is computed efficiently using the chunked kernel of~\citet{yang2024gated}, which partitions the sequence into chunks of size $C{=}64$. Within each chunk, the delta rule corrections are batched into a single matrix operation via the WY representation; across chunks, the state $\mathbf{S}$ is propagated sequentially over $T/C$ steps instead of~$T$, yielding linear-time complexity with high GPU utilization.

\begin{table}[t]
\centering
\label{tab:gdn-dims}
\small
\begin{tabular}{lcc}
\toprule
\textbf{Parameter} & \textbf{Shape} & \textbf{Count} \\
\midrule
$\mathbf{W}^Q$, $\mathbf{W}^K$ & $1536 \times 2048$ & $2 \times 3.15\text{M}$ \\
$\mathbf{W}^V$, $\mathbf{W}^G$, $\mathbf{W}^O$ & $3072 \times 2048$ & $3 \times 6.29\text{M}$ \\
$\mathbf{W}_\alpha$, $\mathbf{W}_\beta$ & $6 \times 2048$ & $2 \times 12.3\text{K}$ \\
$\mathbf{A}_{\log}$, $\Delta_{\text{bias}}$ & $6$ & $12$ \\
Short conv ($Q$, $K$, $V$) & kernel 4 & $\sim 31\text{K}$ \\
\midrule
\textbf{GDN mixer total} & & $\sim 25.2\text{M}$ \\
\bottomrule
\end{tabular}
\caption{GDN parameter dimensions for Llama-3.2-1B ($d = 2048$, $H = 6$, $d_h = 256$, $\alpha = 2$).}
\end{table}

\subsection{Memory-Efficient Long-Context Knowledge Distillation}
\label{sec:memory-efficiency}

Extending knowledge distillation from 2K to 64K context lengths introduces severe memory pressure.
The dominant bottleneck is the \emph{logit tensor}: for sequence length $T$ and vocabulary size $V$, the standard KL divergence loss requires materializing both student and teacher logit matrices of shape $(T, V)$ simultaneously.
At $T{=}65{,}536$ and $V{=}128{,}256$ (Llama-3), each logit tensor consumes approximately 16\,GB in bfloat16---making naive distillation infeasible even on 80\,GB GPUs.
We address this through a progression of increasingly aggressive memory optimizations, summarized in Table~\ref{tab:mem-opts} with training memory reported for some of the optimizations in Table~\ref{tab:train-memory}.

\subsubsection{Fused Linear Cross-Entropy}
\label{sec:fused-ce}

The standard cross-entropy computation first projects hidden states through the LM head $\mathbf{W}_\text{lm} \in \mathbb{R}^{V \times d}$ to produce the full logit matrix $\mathbf{Z} = \mathbf{H}\mathbf{W}_\text{lm}^\top \in \mathbb{R}^{T \times V}$, then applies the softmax and loss in a separate step.
This requires $O(TV)$ memory just for the logit materialization.

\subsubsection{Chunked KL Divergence}
\label{sec:chunked-kl}

The KL divergence $D_\text{KL}(p_s \| p_t)$ between student and teacher logits normally requires both $(T, V)$ log-softmax tensors in memory.
For long contexts, we chunk along the sequence dimension:
\begin{equation}
\label{eq:chunked-kl}
D_\text{KL} = \frac{1}{T} \sum_{i=0}^{\lceil T/C \rceil - 1} \sum_{j=iC}^{\min((i+1)C, T)-1}
D_\text{KL}\!\bigl(\text{softmax}(\mathbf{z}_j^{(s)}) \| \text{softmax}(\mathbf{z}_j^{(t)})\bigr),
\end{equation}
with $C{=}4{,}096$. Each chunk allocates only a $(C, V)$ softmax slice, reducing peak memory from $2 \times T \times V$ to $2 \times C \times V$.
Intermediate tensors are explicitly freed between chunks.

\subsubsection{Triton-Fused KL Divergence}
\label{sec:fused-kl}

For further efficiency, we leverage a custom Triton kernel from FLA~\citep{yang2024fla} that computes $D_\text{KL}$ entirely within a single fused kernel using online softmax~\citep{milakov2018online}.
The kernel tiles over the vocabulary dimension with block size $B_V$ and maintains running log-sum-exp accumulators, so the full $(T, V)$ softmax matrices are never materialized:
\begin{equation}
\label{eq:fused-kl}
D_\text{KL}^{(j)} = \sum_{b=0}^{\lceil V / B_V \rceil - 1} \left[
  \text{accum}\!\Big(
    \log p_s^{(j)}[b B_V : (b{+}1)B_V],\;
    p_t^{(j)}[b B_V : (b{+}1)B_V]
  \Big)
\right],
\end{equation}
where each token $j$ is processed by one Triton program instance, and the gradient $\partial D_\text{KL} / \partial \mathbf{z}^{(s)}$ is written \emph{in-place} during the forward pass (overwriting the student logit buffer), eliminating the need to save activations for backward.

\subsubsection{Fused Hidden-State KL (Logit-Free Distillation)}
\label{sec:fused-kl-hidden}

At the longest contexts (64K tokens), even chunked approaches are bottlenecked by the need to run the teacher's LM head.
We use a \emph{logit-free} distillation path: the teacher forward pass \textbf{skips the LM head entirely}, returning only the final hidden states $\mathbf{H}^{(t)} \in \mathbb{R}^{T \times d_t}$.
The FLA \texttt{FusedKLDivLoss} then computes KL directly from hidden states and LM head weight matrices:
\begin{equation}
\label{eq:fused-kl-hidden}
\mathcal{L} = D_\text{KL}\!\Big(
  \text{softmax}\!\big(\mathbf{H}^{(s)} \mathbf{W}_\text{lm}^{(s)\top}\big)
  \;\Big\|\;
  \text{softmax}\!\big(\mathbf{H}^{(t)} \mathbf{W}_\text{lm}^{(t)\top}\big)
\Big),
\end{equation}
where the softmax and KL are computed in a tiled fashion inside the Triton kernel \emph{without materializing either logit matrix}.
This eliminates $2 \times T \times V$ elements from GPU memory (approximately 32\,GB at $T{=}64$K).
The teacher's LM head weight $\mathbf{W}_\text{lm}^{(t)}$ is accessed via FSDP's \texttt{summon\_full\_params} to avoid duplicating sharded parameters.

\subsubsection{Teacher Memory Management}
\label{sec:teacher-memory}

To accommodate the teacher model (Llama-3.1-8B) alongside the student \ours models during long-context distillation, we employ several additional strategies:

\begin{itemize}[nosep]
  \item \textbf{Frozen teacher under \texttt{torch.no\_grad}:} The teacher runs in evaluation mode with gradient computation disabled, eliminating all optimizer states, gradient tensors, and backward graph storage for teacher parameters.
  \item \textbf{FSDP full sharding for both models:} Both student and teacher are wrapped with FSDP using \texttt{FULL\_SHARD} strategy, distributing model parameters across all 8 GPUs. Only one shard per GPU is materialized at a time during forward/backward passes.
  \item \textbf{bfloat16 mixed precision:} All activations and parameters are stored in bfloat16, halving memory relative to float32.
  \item \textbf{Batch size reduction:} At 64K+ contexts, per-device batch size is reduced to 1 (from 4 at 8K), trading throughput for memory headroom.
  \item \textbf{Activation checkpointing (optional):} Gradient checkpointing is supported for the decoder stack and can be enabled when activation memory dominates; however, the sub-quadratic memory of linear recurrence layers (Mamba-2/GDN) and the fused loss kernels typically suffice without it.
\end{itemize}

\begin{table}[!t]
\centering
\caption{Memory optimization techniques and their deployment across context lengths. Each technique targets a specific memory bottleneck in the knowledge distillation pipeline.}
\label{tab:mem-opts}
\small
\begin{tabular}{llc}
\toprule
\textbf{Technique} & \textbf{Memory Saved}  & \textbf{Used at} \\
\midrule
Liger Fused Linear CE & Student logits $(T {\times} V)$  & 8K--32K \\
Chunked KL Divergence & Softmax tensors $2(T {\times} V)$  & 64K \\
Triton Fused KL & Softmax + grad $3(T {\times} V)$  & 128K \\
Fused Hidden-State KL & Both logit matrices $2(T {\times} V)$  & 64K \\
FSDP Full Sharding & Model params $\div N_\text{GPU}$ & All \\
Frozen teacher (\texttt{no\_grad}) & Teacher grads + optimizer  & All \\
bf16 mixed precision & $2{\times}$ vs.\ fp32  & All \\
\bottomrule
\end{tabular}
\end{table}

\subsection{LLM Usage}
The authors of this paper used AI tools for polishing text within this paper. The authors take full responsibility for the content within this paper.

\end{document}